\documentclass[lettersize,journal]{IEEEtran}

\usepackage{balance}
\usepackage{amsthm}
\usepackage{amsmath,amsfonts}
\usepackage{algorithmic}
\usepackage{algorithm}
\usepackage{array}
\usepackage{bbold}
\usepackage[caption=false,font=normalsize,labelfont=sf,textfont=sf]{subfig}
\usepackage{textcomp}
\usepackage{stfloats}
\usepackage{url}
\usepackage{verbatim}
\usepackage{graphicx}
\usepackage{cite}
\usepackage{lineno,hyperref}
\modulolinenumbers[5]
\usepackage{algorithmic}
\usepackage[dvipsnames]{xcolor}
\usepackage{mathtools}
\usepackage{enumerate}
\usepackage{booktabs}
\usepackage{color}
\newcommand{\mat}[1]{\boldsymbol{#1}}
\newcommand{\real}{\mathbb{R}}
\newcommand{\norm}[1]{\left\lVert#1\right\rVert}
\newtheorem{theorem}{Theorem}
\usepackage{amsmath}

\newcommand{\blockdiag}[1]{\mathrm{diag}\left(#1\right)}

\begin{document}
\title{A Deep Unrolling  Model with Hybrid Optimization Structure for Hyperspectral Image Deconvolution}
\author{Alexandros Gkillas ~\IEEEmembership{Student Member,~IEEE},
Dimitris Ampeliotis, ~\IEEEmembership{Member,~IEEE},
Kostas Berberidis, ~\IEEEmembership{Senior Member,~IEEE}


 \thanks{Alexandros Gkillas and Kostas Berberidis are with the Department of Computer Engineering and Informatics,  Patras University, Patras,
Greece.}
 \thanks{Dimitris Ampeliotis is with the Department Digital Media and Communication, Ionian University,
Argostoli, Greece}

}

\maketitle

\begin{abstract}

In recent literature there are plenty of works that combine handcrafted and learnable regularizers to solve inverse imaging problems. While this hybrid approach has demonstrated
promising results, the motivation for combining handcrafted
and learnable regularizers remains largely underexplored and
has not been thoroughly studied in the existing literature. This work aims to justify this combination,
by demonstrating that the incorporation of proper handcrafted regularizers alongside learnable regularizers not only significantly
reduces the complexity, in terms of the number of parameters required, of the learnable prior, but also the restoration performance
is notably enhanced. To analyze the impact of this synergy,
we introduce the notion of residual structure, to refer to the
structure of the solution that cannot be modeled by the
handcrafted regularizers per se. Using this concept, we experimentally show that the complexity of the learnable regularizer is directly proportional to the amount of the residual structure. Motivated by these findings, we propose a novel optimization framework for the hyperspectral deconvolution problem, called DeepMix.
Based on the proposed optimization framework, an interpretable model is developed  using the deep unrolling
strategy, which consists of three distinct modules, namely,
a data consistency module, a module that enforces the
effect of the handcrafted regularizers, and a denoising
module due to the learnable regularizer. 
Recognizing the collaborative nature of
these modules, this work proposes a context aware
denoising module designed to sustain the advancements
achieved by the cooperative efforts of the other modules.
This is facilitated through the incorporation of a proper skip
connection, ensuring that essential details and structures
identified by other modules are effectively retained and
not lost during denoising.
Extensive experimental results across simulated and real-world datasets demonstrate that DeepMix is notable for surpassing existing methodologies, offering marked improvements in both image quality and computational efficiency. 


\end{abstract}

\begin{IEEEkeywords}
Hyperspectral images, deep equilibrium, deconvolution, deep unrolling
\end{IEEEkeywords}

\section{Introduction}
\IEEEPARstart{I}{n} recent years, hyperspectral imaging (HSI) has emerged as an essential tool in various fields, including remote sensing, medical science, and autonomous driving \cite{remote_sensing_example_2}. HSI offers detailed spectral information through numerous spectral bands that is particularly useful in these areas  \cite{myICIP}. Nonetheless, during the acquisition stage, hyperspectral images are often subject to degradation, including various blurring effects such as atmospheric turbulence, vibrations of the imaging system
platform, and the lens being out-of-focus  which can significantly deteriorate their performance in many HSI applications \cite{good, astro}. Therefore, the restoration of hyperspectral images constitutes a critical pre-processing step. This requirement stresses the need for efficient and effective deconvolution techniques for HSI \cite{Rasti_review, 9144237}.

From a theoretical perspective, hyperspectral image deconvolution can be classified as an inverse imaging problem \cite{10271326, 10138912, 9404870}. Unlike 2D imaging tasks, it presents unique challenges due to the high dimensionality of the data, where each pixel is represented by a spectral vector rather than a single scalar intensity. To address this complexity, optimization-based approaches have traditionally been employed. These methods incorporate hand-crafted regularization terms in the respective cost functions, such as sparsity or low-rank priors, to guide the solution towards the desired properties \cite{cao1, wltr}. In more detail, the handcrafted regularizers rely on domain knowledge to describe general properties of the data, making them interpretable and capable of generalizing well to unseen scenarios. However, their effectiveness is limited by their ability to capture only general data properties, which are often insufficient for state-of-the-art performance, and by the significant computational burden imposed by the large number of optimization steps required \cite{Rasti_review}.
In response to these limitations, learnable regularizers have emerged as an alternative \cite{ dl_review_imaging}. These methods rely on neural networks trained on suitable datasets to implicitly model the desired properties of the solution \cite{learned_regularizers, sos_pnp}. Unlike handcrafted regularizers, learnable regularizers adapt to the data and can capture complex properties that are difficult to be modeled by a human expert working solely with handcrafted regularization terms.

Recent works have explored this concept further by combining handcrafted and learnable regularizers into a single framework \cite{10004995, 10173648, 10282195, 10114568, myTIP}. These methods typically consist of three main components: a data fidelity term (or data consistency module), a handcrafted regularization module, and a learnable regularization module, which is often implemented as a denoising neural network.  While this hybrid approach has demonstrated promising results, the motivation for combining handcrafted and learnable regularizers, remains largely underexplored and has not been thoroughly studied in the existing literature. In this context, a fundamental question arises: \textit{If learnable regularizers can be trained to model complex structures, why are handcrafted regularizers still necessary?}


This work delves deeper towards answering this question, by demonstrating that incorporating proper handcrafted regularizers alongside learnable regularizers not only significantly reduces the complexity of the learnable prior, but also, by carefully designing their collaboration, the restoration performance is notably enhanced.  To analyze the impact of this synergy, we introduce the notion of \emph{residual structure}, to refer to the characteristics of the solution that cannot be modeled by the handcrafted regularizers per se. Furthermore, we experimentally show that when the residual structure is stronger, as it is the case when some of the handcrafted regularization terms are omitted, the model complexity of the lernable regularizer should be increased, as it needs to capture more structure (information) about the image. Likewise, the incorporation of a proper, ``more complete'', set of handcrafted regularizers leads to a learnable regularizer model with reduced complexity (smaller number of parameters), as the learnable regularizer is required to model less information.

To solve the proposed optimization problem, we adopt the deep unrolling framework \cite{li2021deep, 9721184, rebeca}, where a fixed number of optimization iterations are "unrolled" into a model-based neural network, leading to a highly efficient deep architecture. The derived solution consists of three interpretable modules: the data consistency module, the handcrafted regularization module, and the denoising module. However, following the existing deep unrolling approaches  the collaborative interactions between these modules is often overlooked. As a result, the denoising module unnecessarily addresses parts of the solution already identified by the data consistency and handcrafted regularization modules, which increases its complexity in terms of learnable parameters.

To address this limitation, this study enhances the collaboration among the modules by introducing skip connections into the deep unrolling framework. Specifically, we focus on the denoising module, where existing literature often relies on generic large denoising architectures with millions of learnable parameters that are agnostic to the iterative and collaborative nature of the deep unrolling framework. 
In this work, we recognize that the denoising module does not operate in isolation but functions as part of a larger framework, working in conjunction with other modules to progressively enhance the quality of the restored image. Based on this observation, we propose a \textit{context-aware}  denoiser designed to maintain and build upon the progress achieved by the other two collaborating modules. To accomplish this, we modify the denoising neural network by introducing a carefully designed skip connection. This addition ensures that critical information generated by the preceding modules is effectively retained and utilized, rather than being diminished or lost during the denoising process. As a result, the skip connection directs the learnable regularizer to focus on the residual structure, allowing the denoising module to be more compact in terms of learnable parameters by avoiding redundancy in capturing structures already identified by the preceding modules.

The  \textbf{contributions} of this work, which significantly enhance and extend our previous work in \cite{myIcassp}, are summarized as:
\begin{itemize}
\item A novel optimization problem for HSI deconvolution is proposed, involving the combination of a learnable regularizer (in the form of a deep learning network) with suitable handcrafted regularizers. This synergy not only provides enhanced restoration performance, but, more importantly, the handcrafted regularizers allow for a learnable regularizer that is more compact. This finding is validated experimentally, initiating a discussion on the interplay between learnable and handcrafted regularizers.
\item Based on the proposed  optimization framework, an  interpretable model is developed by using the deep unrolling strategy, which consists of three distinct modules, namely, a data consistency module, a
module that enforces the effect of the handcrafted regularizers, and a  denoising module, in the form of a
neural network.
\item Inside the deep unrolling architecture, we replace the denoising module with a more effective module, called \textit{context aware denoising module}.  
 In contrast to
the typical denoising scenario, where networks deal with noisy
 inputs, the deep unrolling setup involves
a series of interconnected modules that together improve
image quality. Hence, the input to the denoiser in a
deep unrolling context is more structured and carries crucial
information. Recognizing the collaborative nature of these
modules, this work proposes a context aware denoising
network designed to sustain the advancements achieved by the
cooperative efforts of these modules.  This is facilitated via
the incorporation of a proper skip connection, which  improves information flow within the architecture, ensuring that essential details identified by other modules are effectively retained and not lost during denoising.


\item Theoretical convergence analysis and guarantees are provided for the proposed method.
\item Extensive  experiments were conducted for the HSI deconvolution problem, demonstrating that the proposed approach outperforms a wide range of state-of-the-art methods on both simulated and real-world datasets. 

\end{itemize}

The remainder of this paper is organized as follows. In Section \ref{positioning}, the positioning of this work in the literature is given. In the sequel, Section \ref{proble_form} formulates the problem under study. Section \ref{problem_form} derives the proposed explainable deep unrolling model. 
Section \ref{conergence} provides the convergence analysis of the proposed model. Section \ref{Results} presents a series of extensive numerical results. Finally, Section \ref{conclusions} concludes the paper.

\section{Related Work} \label{positioning}
In this section, we discuss how the proposed method differs from state-of-the-art techniques in hyperspectral deconvolution, as well as from other deep unrolling approaches developed for inverse imaging problems.
\subsection{Hyperspectral Deconvolution}
The literature contains many studies that focus on the HSI deconvolution problem, adopting diverse perspectives and assumptions to address the associated challenges. Despite the numerous approaches proposed thus far, several challenges remain open due to the inherent trade-off between restoration accuracy and  computation complexity. In particular, the high-dimensional nature of hyperspectral images makes it difficult to achieve low computational complexity while maintaining high restoration quality \cite{9420707, 3DWiener, 1261329}. In the case where the emphasis is on the restoration accuracy rather on the computational complexity, optimization based solutions that employ proper regularization terms are used \cite{Rasti_review}. Such approaches are typically employed in many inverse problems, as long as suitable regularization terms, called handcrafted regularizers that promote desired characteristics of the solution can be devised.  Among the methods that utilize handcrafted regularizers is the one in \cite{method_3}, that combines spatial and spectral priors in the form of optimization constraints, leading to improved deconvolution results. The study in \cite{good} proposes an optimization-based method with non-negative regularization constraints, utilizing minimum distance and maximum curvature criteria for estimating the regularization parameter. In \cite{tv}, the spatial-spectral joint total variation is employed to model the 3D structure of hyperspectral images. Viewing the problem from a spectral perspective, the works in \cite{cao1, cao2} introduce the dark channel prior for HSI deblurring, along with the $l_0$- and $l_1$-based Total Variation (TV) regularizers. In \cite{wltr}, both spatial non-local self-similarity and spectral correlations were investigated using low-rank tensor priors. 
Optimization-based methods that utilize such handcrafted priors are well justified, they offer clear interpretation, but face some limitations such as the high computational demands and the necessity for manual hyperparameter tuning \cite{Rasti_review}.


On the other hand, deep learning techniques applied to hyperspectral imaging (HSI) problems often function as black-box solutions \cite{8557110}. These methods typically lack interpretability and demand large volumes of data to achieve optimal performance. Furthermore, they face generalization issues \cite{Rasti_review, magazine}. Recent SOTA plug-and-play approaches \cite{sos_pnp, wang_main} aim to combine the strengths of both paradigms—optimization-based methods and deep learning—by incorporating a learned neural network prior into the iterative optimization processes. However, these methods still face challenges, such as the necessity for large number of iterations, as the learnable prior is optimized independently from the  problem at hand.  

The resulting method, so-called DeepMix, 
combines the merits of the optimization-based methods and the deep learning techniques. 
The method offers increased restoration performance, and computational efficiency while retaining a clear understanding of the underlying processes. In this work, the focus is on non-blind HSI deconvolution, which is, in fact, an area of significant research activity, as mentioned in  recent studies \cite{sos_pnp, Rasti_review, myIcassp}. 
 Furthermore, it should be highlighted that the proposed approach exhibits significant restoration performance even in the case where the blurring kernel is not accurately known. In more detail, as shown in the  experimental results presented  in Section \ref{gener} and \ref{real-world}, our method shows promising generalization capabilities in this setting. 
 

\subsection{Deep unrolling}
The literature regarding the use of the deep unrolling strategy in imaging tasks is rich, including numerous studies that aim to solve various image restoration problems by either unrolling a small number of iterations of the optimization algorithms, \cite{du_new2, du_new3, du_new4} or by expressing the unrolling methodology  as a fixed point calculation based on the deep equilibrium modeling \cite{rebeca}. Despite the substantial progress, two key limitations remain largely unaddressed in the literature. 

 First, numerous studies on inverse imaging problems have formulated optimization frameworks that combine a data fidelity term with both handcrafted and learnable regularizers.
For instance, works have utilized low-rank and learnable regularizers for color image processing \cite{10004995}, sparsity and deep priors for hyperspectral unmixing \cite{10173648}, and total variation with neural networks for 2D image restoration \cite{HUO2024191}. Similarly, other studies integrated low-rank, sparse, and deep priors for hyperspectral denoising \cite{10282195} or combined non-local similarity with deep priors for hyperspectral super-resolution \cite{10114568}. However, most of these methods do not explicitly address the motivation behind combining handcrafted and learnable regularizers. Furthermore, the potential advantages of such a hybrid approach—compared to relying solely on a learnable regularizer—remain underexplored in the current literature.

In light of these, this study argues that handcrafted regularizers, while unable to capture all the properties of the data, play a crucial role in modeling important aspects of the solution. Their inclusion not only significantly reduces the complexity of the learnable prior but also enhances restoration performance by allowing the learnable prior to focus on capturing more complex properties of the data that cannot easily be modeled by handcrafted regularizers. To analyze this interplay, we introduce the notion of \emph{residual structure}, which refers to the characteristics of the solution that remain unmodeled by the handcrafted regularizers. The greater the residual structure—for example, in the case  when some handcrafted regularization terms are omitted—the higher the complexity (e.g., number of parameters) required of the learnable regularizer to model these additional details. Conversely, the inclusion of a well-chosen  set of handcrafted regularizers reduces this residual structure, enabling a simpler, more efficient learnable model with relaxed complexity (e.g., a smaller neural network), while maintaining superior performance.

Secondly, while the learned regularizer's effect is typically enforced through a trained denoising neural network, deep unrolling approaches often use generic denoising architectures for this task \cite{rebeca, du_new4, shlezinger2023model}. However, the denoising task in deep unrolling differs significantly from classical denoising. Unlike conventional scenarios, where networks process noisy and unstructured inputs, deep unrolling involves progressively denoised inputs derived from interconnected modules, which collaboratively improve image quality. To address this, we propose a context-aware denoising network that incorporates skip connections, ensuring that information flows efficiently through the architecture and that crucial details captured by other modules are preserved during the denoising process.
Moreover, the skip connection allows the learnable regularizer to focus on the residual structure that the other modules cannot model. This design prevents redundancy in the size of the denoiser, ensuring that the denoising module does not expend resources rediscovering already identified structures.

\section{Problem Formulation}
\label{proble_form}

 Let us denote a hyperspectral image as $\mathbf{X} \in \mathbb{R}^{M \times N \times d}$, where $M$ and  $N$ represent the spatial dimensions of the image, and $d$ represents the number of spectral bands or channels.
The corresponding degraded or blurred image is represented as $\mathbf{Y} \in \mathbb{R}^{M \times N \times d}$. 
 Based on the linear degradation model \cite{linear_noise, wang_main} the $i$-th spectral band of the degraded image $\mat{Y}_i \in \real^{M \times N}$ is given by the following relation
\begin{equation}
    \mat{Y}_i = \mat{H_i} \star \mat{X}_i + \mat{W}_i, \quad i=1,\ldots,d,
    \label{eq:blur0}
\end{equation}
where $\star$ represents the convolution operator, $\mat{H_i}$ is the blurring kernel of the $i$-th spectral band and $\mat{W}_i$ is a noise term.

Given the degraded hyperspectral image $\mat{Y}=\{\mat{Y}_i\}_{i=1}^d$, the objective is to recover the clean image $\mat{X}$. Considering that the signals
exhibit string spatial and spectral dependencies, this a-priori information can be utilized to derive a novel optimization problem. To this end we explore the idea of combining handcrafted regularizers  with learnable regularizers that aim to capture the \textbf{residual structure} of the data which cannot be modeled from the handcrafted regularizers.  Thus, we propose the following optimization problem
\begin{align}
\underset{\mat{X}}{\arg\min}\,\,\, \frac{1}{2} \sum_{i=1}^d\norm{\mat{Y}_i - \mat{H_i} \star \mat{X}_i}_F^2  + \lambda \mathcal{R}(\mat{X})\ \nonumber \\
+ \mu_v\norm{\nabla_v \mat{X}}_F^2  + \mu_h\norm{\nabla_h \mat{X}}_F^2 + \mu_s\norm{\nabla_s \mat{X}}_F^2 , 
\label{eq:main_problem}
\end{align}
which consists of a data consistency module that enforce the degradation model in relation (\ref{eq:blur0}) and the handcrafted regularizers $\nabla_v$, $\nabla_h$, and $\nabla_s$, which  represent linear operators (matrices)  for the horizontal, vertical, and spectral gradients of $\mat{X}$, respectively. By incorporating these gradients, our approach enforces smoothness priors into the solution, preserving a strong property of the hyperspectral images. However, handcrafted regularizers alone cannot model the entire complexity of the
underlying structure of the data,
To complement the handcrafted regularizers, we incorporate a learnable regularizer $\mathcal{R}(\cdot)$, which models the \textbf{residual structure}, defined as the structure of the solution  that handcrafted priors cannot capture. By including the handcrafted terms, the residual structure is reduced, as they already model much of the spatial and spectral smoothness inherent to hyperspectral signals. This reduction allows the learnable regularizer to focus solely on the remaining, more intricate properties of the data, such as higher-order dependencies or localized variations. Consequently, the learnable prior can be designed with fewer parameters, resulting in a simpler and more efficient model, without compromising restoration accuracy. 
Finally, $\lambda, \mu_v, \mu_h, \mu_s$ are the regularization parameters. 

\section{Proposed Optimization based Model} \label{problem_form}
In this section, first we solve the proposed  optimization problem using the HQS methodology (section \ref{hqs_section}), next we express the derived steps of the algorithm as individual modules (section \ref{modules_section}) and finally we use these modules to design an interpretable deep unrolling model (section \ref{DEQ}). 
\subsection{Iterative Optimization algorithm} \label{hqs_section}
To effectively address problem (\ref{eq:main_problem}), 
the HQS technique  is used, given its straightforward implementation and proven convergence capability in a variety of applications \cite{zhang2020deep}. By incorporating some auxiliary variables i.e.,  $\mat{Z}, \mat{V}_1, \mat{V}_2, \mat{V}_3 \in \real^{M \times N \times d}$, equation (\ref{eq:main_problem}) can be reformulated equivalently as 
\begin{align}
   \mat{X} = &\underset{\mat{X}}{\arg\min}\,\,\, \frac{1}{2} \sum_{i=1}^d\norm{\mat{Y}_i - \mat{H_i} \star \mat{X}_i}_F^2  + \lambda \mathcal{R}(\mat{Z}) \nonumber \\
&+ \mu_v\norm{\nabla_v \mat{V}_1}_F^2  + \mu_h\norm{\nabla_h \mat{V}_2}_F^2 + \mu_s\norm{\nabla_s \mat{V}_3}_F^2 \nonumber\\
   &s.t. \quad \mat{Z}-\mat{X}=0, \mat{V}_1-\mat{X}=0, \nonumber\\ 
   &\quad \quad \, \mat{V}_2-\mat{X}=0, \mat{V}_3-\mat{X}=0. 
\end{align}
The HQS aims to minimize the following loss function
\begin{align}
    \mathcal{L} = \frac{1}{2} \sum_{i=1}^d\norm{\mat{Y}_i - \mat{H_i} \star \mat{X}_i}_F^2  + \lambda \mathcal{R}(\mat{Z}) 
   + \frac{b_1}{2}\norm{\mat{Z} -\mat{X}}_F^2 \nonumber\\
   + \mu_v\norm{\nabla_v \mat{V}_1}_F^2  + \mu_h\norm{\nabla_h \mat{V}_2}_F^2 + \mu_s\norm{\nabla_s \mat{V}_3}_F^2 \nonumber\\
   + \frac{b_2}{2}\norm{\mat{V}_1 -\mat{X}}_F^2 + \frac{b_3}{2}\norm{\mat{V}_2 -\mat{X}}_F^2 +  \frac{b_4}{2}\norm{\mat{V}_3 -\mat{X}}_F^2 \label{eq:Lagrangian}
\end{align}
where $b_1,b_2, b_3, b_4$ stand for the penalty parameters.
Based on (\ref{eq:Lagrangian}), 
a sequence  of iterative sub-problems is derived
\begin{subequations}
\small
\begin{align}
    \mat{X}^{(k+1)} = &\underset{\mat{X}}{\arg\min}\,\frac{1}{2} \sum_{i=1}^d\norm{\mat{Y}_i - \mat{H_c} \star \mat{X}_i}_F^2 \nonumber \\ 
 &+ \frac{b_1}{2}\norm{\mat{Z}^{(k)} -\mat{X}}_F^2 
 + \frac{b_2}{2}\norm{\mat{V}_1^{(k)} -\mat{X}}_F^2 \nonumber \\ 
 &+ \frac{b_3} {2}\norm{\mat{V}_2^{(k)} -\mat{X}}_F^2 + \frac{b_4}{2}\norm{\mat{V}_3^{(k)} -\mat{X}}_F^2
     \label{eq:updateX}\\     
   \mat{Z}^{(k+1)} = &\underset{\mat{Z}}{\arg\min}\, \lambda \mathcal{R}(\mat{Z}) 
   + \frac{b_1}{2}\norm{\mat{Z} -\mat{X}^{(k+1)}}_F^2\  \label{eq:updateZ}\\
   \mat{V}_1^{(k+1)} = &\underset{\mat{V}_1}{\arg\min}\,  \mu_v\norm{\nabla_v \mat{V}_1}_F^2 
   + \frac{b_2}{2}\norm{\mat{V}_1 -\mat{X}^{(k+1)}}_F^2   \label{eq:updateV1}
\\
   \mat{V}_2^{(k+1)} = &\underset{\mat{V}_2}{\arg\min}\,  \mu_h\norm{\nabla_h \mat{V}_2}_F^2 
   + \frac{b_3}{2}\norm{\mat{V}_2 -\mat{X}^{(k+1)}}_F^2   \label{eq:updateV2}
\\
   \mat{V}_3^{(k+1)} = &\underset{\mat{V}_3}{\arg\min}\,  \mu_s\norm{\nabla_s \mat{V}_3}_F^2 
   + \frac{b_4}{2}\norm{\mat{V}_3 -\mat{X}^{(k+1)}}_F^2   \label{eq:updateV3}
\end{align}
\normalsize
\end{subequations}

\subsection{Consequent Modules}\label{modules_section}

\subsubsection{\textbf{Data Consistency Module - Sub-problem (\ref{eq:updateX}) }}
The Data Consistency module aims to derive a clearer hyperspectral image by imposing a degradation physical model that corresponds to the  problem of deconvolution.
Utilizing the fast Fourier transform (FFT), sub-problem (\ref{eq:updateX}) can be written into a more efficient form into the Fourier domain,
\begin{align}
\small
    \mat{\tilde{X}}^{(k+1)} = \underset{\mat{\tilde{X}}}{\arg\min}&\frac{1}{2} \norm{\mat{\tilde{Y}} - \mat{\tilde{H}} 	\odot \mat{\tilde{X}}}_F^2  
    + \frac{b_1}{2}\norm{\mat{\tilde{Z}}^{(k)} -\mat{\tilde{X}}}_F^2  \nonumber \\
    & +\frac{b_2}{2}\norm{\mat{\tilde{V}}_1^{(k)} -\mat{\tilde{X}}}_F^2     + \frac{b_3}{2}\norm{\mat{\tilde{V}}_2^{(k)} -\mat{\tilde{X}}}_F^2 \nonumber \\
    &+ \frac{b_4}{2}\norm{\mat{\tilde{V}}_3^{(k)} -\mat{\tilde{X}}}_F^2
    \label{eq:updateX_fourier}
\normalsize
\end{align}
where $\tilde{\mat{Y}}=\{\mathcal{F}(\mat{Y}_i)\}_{i=1}^d$, $\tilde{\mat{H}}=\{\mathcal{F}(\mat{H}_i)\}_{i=1}^d$, $\tilde{\mat{X}}=\{\mathcal{F}(\mat{X}_i)\}_{i=1}^d$, $\tilde{\mat{Z}}=\{\mathcal{F}(\mat{Z}_i)\}_{i=1}^d$, $\tilde{\mat{V}_1}=\{\mathcal{F}(\mat{V_1}_i)\}_{i=1}^d$, $\tilde{\mat{V}_2}=\{\mathcal{F}(\mat{V_2}_i)\}_{i=1}^d$ and $\tilde{\mat{V}_3}=\{\mathcal{F}(\mat{V_3}_i)\}_{i=1}^d$   stand for the concatenation of the discrete 2D Fourier transforms for each spectral band of the respective spatial domain signals (i.e.,  $\mat{Y}_i, \mat{H}_i, \mat{X}_i, \mat{Z}_i, \mat{V_1}_i, \mat{V_2}_i, \mat{V_3}_i $). Here, $\mathcal{F}(\cdot)$ represents the 2D Fourier transform.
The solution is given by
\begin{align}
    \mat{\tilde{X}}^{(k+1)} = \left( \mat{\tilde{H}} \odot \mat{\tilde{H}} + (b_1 + b_2 +b_3 + b_4)\mathbb{1}\right)^{-1}\odot \nonumber \\
    (\mat{\tilde{H}} \odot \mat{\tilde{Y}} + b_1\mat{\tilde{Z}}^{(k)} + b_2\mat{\tilde{V}}_1^{(k)} + b_3\mat{\tilde{V}_2}^{(k)} + b_4\mat{\tilde{V}_3}^{(k)})\ .
    \label{eq:x_solution}
\end{align}
Since this solution incorporates the blurring operator, it can efficiently enforce data fidelity while embedding valuable structural properties from the degradation model into the restored image.

\subsubsection{\textbf{Smooth Gradients Prior module - Sub-problems (\ref{eq:updateV1}, \ref{eq:updateV2}, \ref{eq:updateV3} )}}
Concerning these sub-problems  (\ref{eq:updateV1}, \ref{eq:updateV2}, \ref{eq:updateV3}), we derive the following closed-form solutions
\begin{subequations}
\begin{align}
\small
   &\mat{V}_1^{(k+1)} = (\mat{I} + \frac{b_2}{\mu_v}\nabla^T_v\nabla_v)^{-1}\mat{X}^{(k+1)} \label{eq:solutionV1}
\\
   &\mat{V}_2^{(k+1)} = (\mat{I} + \frac{b_3}{\mu_h}\nabla^T_h\nabla_h)^{-1}\mat{X}^{(k+1)} \label{eq:solutionV2}
\\
   &\mat{V}_3^{(k+1)} = (\mat{I} + \frac{b_4}{\mu_s}\nabla^T_s\nabla_s)^{-1}\mat{X}^{(k+1)} \label{eq:solutionV3}
\end{align}
\normalsize
\end{subequations} 
Note that the primary goal of the above solutions is to promote smoothness across the vertical, horizontal and spectral dimensions of the HSI image by surpassing discontinuities due to the blurring effect and the noise. In light of this observation, the above solutions can be replaced by learnable convolution layers consisting of kernels with specific structure to approximate the first-order difference operations of the gradients $\nabla_v, \nabla_h, \nabla_s$. Thus, the above equations are transformed as 
\begin{subequations}
\begin{align}
\small
   &\mat{V}_1^{(k+1)} = \mat{P}_1 * \mat{X}^{(k+1)} \label{eq:learnV1}
\\
   &\mat{V}_2^{(k+1)} = \mat{P}_2 * \mat{X}^{(k+1)} \label{eq:learnV2}
\\
   &\mat{V}_3^{(k+1)} = \mat{P}_3 * \mat{X}^{(k+1)} \label{eq:learnV3}
\end{align}
\normalsize
\label{eq:smooth}
\end{subequations} 
where $P_1$ denotes a conv. layer with a kernel $(L \times 1)$ to approximate the vertical gradient calculation and  replace the
matrix dependent on $\nabla_v$ , $P_2$ convolution layer with a kernel $(1 \times L)$ to approximate the horizontal gradient calculation and replace the
matrix dependent on $\nabla_h$  and $P_3$ is a convolution layer with a kernel $(L \times 1 \times 1)$ to approximate the spectral gradient calculation and replace the
matrix dependent on $\nabla_s$. Note that $L$ is selected to be relatively small and equal to $L=4$ in order to capture more generic structures such as the smoothness of neighboring pixels. The above kernels can be optimized  using the deep unrolling strategy (Section \ref{DEQ}). 

\textbf{Residual Structure:}
By incorporating these gradients, our approach enforces smoothness priors in the solution, capturing a key property of hyperspectral images: spatial and spectral continuity. However, handcrafted regularizers are inherently limited in their ability to fully capture the complexity of the underlying data structure. To address this limitation, we introduce a learnable denoiser in the following section, designed to enforce the residual structure i.e., the aspects of the solution that the handcrafted priors cannot adequately represent. Importantly, the inclusion of handcrafted priors reduces the residual structure, allowing the learnable regularizer to focus on refining the remaining more complex structures. This reduction enables a simpler, more efficient model with fewer parameters, while maintaining high restoration accuracy.

\subsubsection{\textbf{Context aware denoising module - Sub-problem (\ref{eq:updateZ})}}
Considering sub-problem (\ref{eq:updateZ}), it can be expressed as 
\begin{align}
\small
    \mat{Z}^{(k+1)} = 
    \underset{\mat{Z}}{\arg\min} \frac{1}{2(\sqrt{\lambda/b_1})^2}
    \norm{\mat{Z}-\mat{X}^{(k+1)}}_F^2 
    + \mathcal{R}(\mat{Z})
    \label{eq:proximal_Z}
    \normalsize
\end{align}
 Equation (\ref{eq:proximal_Z}) corresponds to a Gaussian denoiser from a Bayesian perspective \cite{Zhangcvpr}. In view of this, a neural network can be employed as a denoiser, to capture the \textbf{residual structure} of the hyperspectral image which cannot be modeled from the handcrafted regularizers in equations (\ref{eq:learnV1}, \ref{eq:learnV2}, \ref{eq:learnV3}). Consequently, equation (\ref{eq:proximal_Z}) can be formulated as:
\begin{equation}
        \mat{Z}^{(k+1)} = f_\theta(\mat{X}^{(k+1)})
        \label{eq:denoiser_z}
\end{equation}
where $f_\theta(\cdot)$ denotes the neural network and $\theta$ its  weights. 

In this work we claim that the denoising network described by equation (\ref{eq:denoiser_z}) represents a significant departure from traditional denoising networks. Unlike conventional models that primarily aim to clean noisy and unstructured data, this network is designed with a fundamentally difference.  Actually, 
this denoising network does not act independently  but in collaboration with the other two modules i.e., the data-consistency and smooth prior modules that also progressively restore the quality of the image. In light of this,  we formulate a \emph{context aware} denoising network that tries to maintain the progress made by the other, collaborating modules. We achieve this behavior by modifying the denoising neural network to also include a proper skip connection. \textit{The skip connection is crucial to the proposed architecture as the input of the denoiser derives from the solution of  sub-problems (\ref{eq:updateX_fourier} and \ref{eq:smooth}), thus maintaining important information from the data consistency and smooth prior modules.} Note that these two modules impose the
degradation constraint and the structure from the handcrafted regularizers, respectively.

In more detail, concerning the network architecture, we employ a simple CNN-based deep learning model that utilizes 3D convolution. 
For the denoising part of the model, the process starts with the input going through a 3D convolution layer, followed by a ReLU activation function. Following this, the network is composed of N blocks of 3D convolution, each consisting of a 3D convolutional layer, batch normalization, and ReLU activation. The final stage includes a proper skip connection that combines the output from the last layer with the input data. Figure \ref{fig:3d-cnn} exemplifies the denoiser architecture.


\subsubsection{\textbf{Overall Iterative Algorithm}} Combining the solutions of each derived module, the final iterative  solutions are
\begin{subequations}\label{eq:HQS_final}
\begin{align}
&\mat{V}_1^{(k+1)} = P_1*\mat{X}^{(k)}\label{eq:gradient_1}\\
&\mat{V}_2^{(k+1)}  = P_2*\mat{X}^{(k)}\label{eq:gradient_2}\\
&\mat{V}_3^{(k+1)} = P_3*\mat{X}^{(k)}\label{eq:gradient_3}\\
&\mat{\tilde{X}}^{(k+1)} = \left( \mat{\tilde{H}} \odot \mat{\tilde{H}} + \alpha_1\mathbb{1}\right)^{-1}\odot \nonumber \\
    &(\mat{\tilde{H}} \odot \mat{\tilde{Y}} + b_1\mat{\tilde{Z}}^{(k)} + b_2\mat{\tilde{V}}_1^{(k+1)} + b_3\mat{\tilde{V}_2}^{(k+1)} + b_4\mat{\tilde{V}_3}^{(k+1)}) 
\label{eq:x}\\
&\mat{Z}^{(k+1)}  = f_\theta(\mat{X}^{(k+1)})\label{eq:denoiser}
\end{align}
\end{subequations} 
where $\mat{Z}, \mat{V}_1, \mat{V}_2, \mat{V}_3, \mat{X}$ denote the signals in the spatial domain, $\tilde{\mat{Z}}, \tilde{\mat{V}}_1, \tilde{\mat{V}}_2, \tilde{\mat{V}}_3, \tilde{\mat{X}}$ are the signal in the Fourier domain and   $\alpha_1 = (b_1 + b_2 +b_3 + b_4)$.
Based on the final iteration map in (\ref{eq:HQS_final}), the above equations can be transformed into an explainable deep learning architecture utilizing the deep unrolling strategy. 

 \subsection{Proposed Interpretable Deep Unrolling model} \label{DEQ}
The proposed iteration algorithm contains learnable parameters that can be optimized via backpropagation i.e., the weights of the denoiser $f_\theta(\cdot)$, the convolution kernels $P_1, P_2, P_3$ that promote smoothness properties across the dimensions of the HSI data and the penalty parameters $\alpha_1, b_1, b_2, b_3, b_4$.  Our aim is to develop an explainable deep network based on the proposed well-justified optimization solver (\ref{eq:HQS_final}).

Following the deep unrolling strategy, the iterations of the HQS solver (\ref{eq:HQS_final}) can be unrolled to establish a highly interpretable network. Each iteration of the proposed solver corresponds to a distinct layer within this unrolling architecture. To further improve the effectiveness  of this model, we express the unrolling iterations as a fixed point equation problem inspired by the Deep Equilibrium models \cite{DEQ, rebeca}.
Specifically, let us represent the proposed HQS solver in (\ref{eq:HQS_final}), consisting of the $5$ update rules as follows
\begin{equation}
 \mat{X}^{(k+1)} = g_\theta(\mat{X}^{(k)};\mat{Y})\ .
 \label{eq:fixed_point_iterations}
\end{equation}
We assume that equation (\ref{eq:fixed_point_iterations}) constructs a deep learning model with an infinite number of layers. By applying the iteration in (\ref{eq:fixed_point_iterations}) infinitely, the output will converge to a fixed point. Consequently, it will satisfy the following fixed point relation
\begin{equation}
    \mat{X}^\star = g_\theta(\mat{X}^\star;\mat{Y})\ .
    \label{eq:fixed_point}
\end{equation}
As a result, using the fixed point theory, the computation of the outputs for the proposed approach can be accelerated, as we can see in Section \ref{forward_pass}. To train the parameters of the proposed infinite-depth architecture an end-to-end training approach can be employed.  Overall, the proposed architecture can be characterized explainable as  each layer consists of three interpretable modules i.e., the data consistency module  (\ref{eq:x}), the context aware denoising module in (\ref{eq:denoiser}) and the smooth gradient priors module consisting of equations (\ref{eq:gradient_1}, \ref{eq:gradient_2}, \ref{eq:gradient_3}). 


\subsection{Implementation Details}

\subsubsection{Fixed point Calculation: Forward and Backward pass} \label{forward_pass}
With the iteration map $g_\theta(\cdot, Y)$ effectively designed, two key challenges emerge. The first challenge consists in computing a fixed point during the forward pass, given a noisy blurred image $Y$ and the neural network denoiser weights $\theta$. The second challenge  is concerned with efficiently obtaining network parameters during the training process.


\textit{Forward pass - Calculating fixed points:}
In the proposed DeepMix, during both training and testing phases, a large number of fixed-point iterations must be executed based on the iteration map in equation (\ref{eq:HQS_final}) to find the fixed point
\begin{equation}
    X^{\star} = g_\theta(X^{\star}, Y)
    \label{eq:fixed_point_vector}
\end{equation}
Hence, the fixed-point iteration can be performed
\begin{equation}
    X^{(k+1)} = g_\theta(X^{(k)}, Y)
\end{equation}
This iterative process continues until convergence, i.e., when $X^{(k+1)}$ and $X^{(k)}$ differ minimally. To reduce computational load, we employ Anderson acceleration \cite{aderson} for faster convergence. 


\textit{Backward pass - Gradient computation:}\label{deq_training}
During the training stage of the proposed DeepMix approach, we utilize the implicit backpropagation to determine the optimal weights $\theta$. As described in \cite{DEQ} and \cite{tutorial}, the objective is to train the proposed model efficiently, avoiding the need  for back-propagation across numerous iterations to reach the fixed point. Instead of  backpropagating through numerous fixed-point iterations, the implicit function theorem is employed to simplify the gradient computation process \cite{DEQ}. This enables a more efficient update of the model's weights during training.

\subsubsection{Loss Function}
Concerning the learnable parameters that consists of the  iteration map in (\ref{eq:fixed_point_iterations}), the DeepMix model can be trained end-to-end using a loss function, as
\begin{equation}
\sum_{p=1}^{P} \norm{g_\theta(\mat{X}^{\star, p},\mat{Y}^{p}) - \mat{X}^{p}}_1,
\end{equation} 
where ${\mat{X}^{p}, \mat{Y}^{p}}$ represent $P$ appropriate training pairs of clean hyperspectral images and their corresponding noisy and blurred images. Additionally, $g_\theta(\mat{X}^{\star, p};\mat{Y}^{p}) = \mat{X}^{\star, p}$ is the output of the proposed DeepMix model given the degraded HSI $\mat{Y}^{p}$. In our implementation the $l_1$ loss function was employed as provided the best restoration results.

\section{Convergence Analysis}\label{conergence}

This section presents a convergence analysis for the proposed model. Recall that some operations of the proposed iteration map in (\ref{eq:HQS_final}) e.g., equation (\ref{eq:x}) are expressed in the frequency domain in order to reduce the computational complexity of the iteration map.  However, for simplicity, we derive an equivalent iteration map that does not involve such operations in the Fourier domain, in a mathematically equivalent manner.  Therefore the final outcomes of the analysis remain valid independently of the domain in which the implementation of the recursions is performed.

\subsection{Equivalent Problem Formulation}
Following the problem formulation of study \cite{sos_pnp, method_2}, the hyperspectral degradation model in (\ref{eq:blur0}) can be expressed as
\begin{equation}
y_i = \mathcal{H}_i x_i + w_i
\end{equation}
where $y_i \in \real^{MN \times 1}$ and $x\in \real^{MN\times 1}$ are the vectorized versions of the i-th spectral bands  $\mat{Y_i} \in \real^{M \times N }$ and $\mat{X_i} \in \real^{M \times N}$, respectively. $\mathcal{H}_i$ is a block Toeplitz matrix of size $MN \times MN$ with $M\times N$ Toeplitz blocks (see also study \cite{sos_pnp}). $\mathcal{H}_i$ can be rewritten as a block circulant matrix with circulant
blocks, denoted as circulant-block-circulant (CBC), ensuring the equivalence with the proposed designed optimization problem. Based on study  \cite{sos_pnp}) the above degradation model is 
\begin{equation}
y = \mathcal{H} x + w
\end{equation}
where $\mathcal{H}$ denotes a block diagonal matrix  of size $MNd \times MNd$, defined as follows
\begin{equation}
\mathcal{H} = \blockdiag{\mathcal{H}_1, \mathcal{H}_2, \dots, \mathcal{H}_d}
\end{equation}
Additionally,  $y \in \real^{MNd \times 1}$ and $x\in \real^{MNd\times 1}$ are the vectorized versions of the images $\mat{Y} \in \real^{M \times N \times d }$ and $\mat{X} \in \real^{M \times N \times d}$, respectively.

 Using the HQS method and introducing  auxiliary variables $z, v_1, v_2, v_3$, the  HSI deconvolution problem is formulated as
\begin{align}
   &\underset{x}{\arg\min}\,\,\, \frac{1}{2} \norm{y - \mathcal{H} x}_F^2  + \lambda \mathcal{R}(z) \nonumber \\
&+ \mu_v\norm{\nabla_v v_1}_F^2  + \mu_h\norm{\nabla_h v_2}_F^2 + \mu_s\norm{\nabla_s v_3}_F^2 \nonumber\\
   &s.t. \quad z-x=0, v_1-x=0, 
    v_2-x=0, v_3-x=0. \nonumber
\end{align}

The loss function that HQS aims to minimize is
\begin{align}
    \mathcal{L} = \frac{1}{2} \norm{y - \mathcal{H}  x}_F^2  + \lambda \mathcal{R}(z) 
   + \frac{b_1}{2}\norm{z -x}_F^2 \nonumber\\
   + \mu_v\norm{\nabla_v v_1}_F^2  + \mu_h\norm{\nabla_h v_2}_F^2 + \mu_s\norm{\nabla_s v_3}_F^2 \nonumber\\
   + \frac{b_2}{2}\norm{v_1 -x}_F^2 + \frac{b_3}{2}\norm{v_2 -x}_F^2 +  \frac{b_4}{2}\norm{v_3 -x}_F^2 \label{eq:Lagrangian23}
\end{align}
The corresponding solutions of the HQS method are 
\begin{subequations}
\begin{align}
&v_1^{(k)} = P_1*x^{(k)}, v_2^{(k)}  = P_2*x^{(k)}, v_3^{(k)} = P_3*x^{(k)}\nonumber\\
&z^{(k)}  = f_\theta(x^{(k)}) \nonumber\\
&x^{(k+1)} = (\mathcal{H}^T  \mathcal{H} + \alpha \mat{I})^{-1} \nonumber \\
    &(\mathcal{H}^T y + b_1 z^{(k)} + b_2 v_1^{(k)} + b_3 v_2^{(k)} + b_4 v_3^{(k)}) 
\label{eq:HQS_final_spatial}
\end{align}
\end{subequations} 
which form the following compact iteration map $h_\theta(., y)$ 
\begin{align}
x^{(k+1)}=&(\mathcal{H}^T  \mathcal{H} + \alpha \mat{I})^{-1}
(\mathcal{H}^T y + b f_\theta(x^{(k)}) \nonumber\\
&+ b_2 P_1 * x^{(k)} + b_3 P_2 * x^{(k)} + b_4 P_3 * x^{(k)}  )
\label{eq:analysis_map}
\end{align}
where $f_\theta(.)$ is the neural network that act as prior for the considered data, $\alpha = b_1 +b_2 +b_3+b_4$ and $P_1, P_2, P_3$ are the convolution layers, similar to equations (\ref{eq:gradient_1}, \ref{eq:gradient_3}, \ref{eq:gradient_3}).
Note that the only difference between the above iteration map and the one in the equation \ref{eq:HQS_final} is that the update of the variable $x$ is performed in the spatial domain. 


\subsection{Convergence Guarantees} \label{theorem}
Given the iteration map in 
\ref{eq:analysis_map}, i.e., $h_\theta(., y): \real^d \rightarrow \real^d$, we aim to provide guarantees that the fixed point equation $x^{(k+1)} = h_\theta(x^{(k)}, y)$ converges to a unique fixed point $x^\star$ as $k \rightarrow \infty$. The classical fixed-point theory \cite{banach} states that the iterative process converges to a unique fixed-point when the iteration map, denoted by $h_\theta(\cdot; y)$, is contractive. Note that an iteration map is contractive if there is a constant $c$ in the range $0 \leq c < 1$, so that  the condition $||h_\theta(x_1; y) - h_\theta(x_2; y)|| \leq c||x_1 - x_2||$ is satisfied for all $x_1, x_2$.  

We assume that the neural network-denoiser, represented by $f_\theta(\cdot)$, is $\epsilon_1$-Lipschitz continuous. This means that there exists a $\epsilon_1 > 0$ such that for all $x_1, x_2$, the following holds
\begin{equation}
    ||f_\theta(x_1) - f_\theta(x_2)|| \leq \epsilon_1 ||x_1 - x_2||.
    \label{eq:Lips_deoniser}
\end{equation}
Also, the convolutional layers $P_1, P_2, P_3$ are $\epsilon_2, \epsilon_3, \epsilon_4 $-Lipschitz continuous.
The above constraints can be easily satisfied by utilizing spectral normalization, which guarantees that the neural network will be e-Lipschitz continuous \cite{miyato2018spectral}.

\begin{theorem}
    Let $f_\theta(\cdot)$ be a neural network denoiser that is $\epsilon_1$-Lipschitz continuous, the conv. layers $P_1, P_2, P_3$ be $\epsilon_2, \epsilon_3, \epsilon_4 $-Lipschitz continuous and let $L =  \lambda_{\mathcal{H}^T\mathcal{H}, min}$ be the minimum eigenvalue of $\mathcal{H}^T\mathcal{H}$. Iteration map $h_\theta(., y)$ (\ref{eq:analysis_map}) satisfies
    \begin{align}
     ||h_\theta(x_1, y) - &h_\theta(x_1, y)|| \leq  \\
     &\frac{(b_1 \epsilon_1 + b_2 \epsilon_2 + b_3 \epsilon_3 + b_4 \epsilon_4 )}{(\alpha + L)^2 } ||x_1-x_2||
    \end{align}
for all $x_1, x_2 \in \real^d$. The iteration map is contractive if $\frac{(b_1 \epsilon_1 + b_2 \epsilon_2 + b_3 \epsilon_3 + b_4 \epsilon_4 )}{(\alpha + L)^2 } <1$, in which case the proposed DeepMix converges. Note that $\alpha = b_1 +b_2 +b_3+b_4$. 
\label{theo}
\end{theorem}

\begin{proof}
Let us consider the  designed iteration map $h_\theta(x, y)$ in equation (\ref{eq:analysis_map}). Its Jacobian with respect to $x \in \mathbb{R}^d$, denoted as $\partial_x h_\theta(x; y)$, is defined as follows
    \begin{align}
    \small
        &\partial_x h_\theta(x; y) =  (\mathcal{H}^T \mathcal{H} + \alpha \mat{I})^{-1} 
        ( b_1 \partial_x f_\theta(x) \nonumber \\
        &+ b_2\partial_x P_1*x + b_3\partial_x P_2*x + b_4\partial_x P_3*x)
        \in \real^{d \times d}
        \label{eq:Jacobian2}
        \normalsize
    \end{align}
where $\partial_x f_\theta(x)$ denotes the Jacobian of $f_\theta(\cdot)$: $\real^d \rightarrow \real^d$ w.r.t x, and $\partial_x P_1, \partial_x P_2, \partial_x P_3$ denotes the Jacobian of $P_1, P_2, P_3$: $\real^d \rightarrow \real^d$ w.r.t x.  To prove that $h_\theta(\cdot; y)$ is a contractive mapping, we need to show that $||\partial_x h_\theta(x; y)|| < 1$ holds for every $x \in \real^d$. Here, $|| \cdot ||$ represents the spectral norm. In addition, to simplify the notation,  the matrix $\mat{B} =  (\mathcal{H} \mathcal{H} + \alpha \mat{I})^{-1}$ is defined. Based on equation (\ref{eq:Jacobian2}), the spectral norm of the Jacobian of the iteration map is given by
\begin{align}
    ||\partial_x h_\theta(x; y)&|| = ||\mat{B} ( b_1 \partial_x f_\theta(x) + b_2\partial_x P_1*x + b_3\partial_x P_2*x \nonumber\\
    &\quad \quad \quad \quad \quad \quad \quad \quad + b_4\partial_x P_3*x)|| \nonumber \\
                                  &\leq||\mat{B}||\, || b_1 \partial_x f_\theta(x) + b_2\partial_x P_1*x + b_3\partial_x P_2*x \nonumber\\
    &\quad \quad \quad \quad \quad \quad \quad \quad + b_4\partial_x P_3*x|| \nonumber \\ \nonumber \\
    &\leq||\mat{B}||( || b_1 \partial_x f_\theta(x)|| + ||b_2\partial_x P_1*x||  \nonumber\\ 
    &\quad \quad \quad \quad \quad  + ||b_3\partial_x P_2*x|| + ||b_4\partial_x P_3*x||) \nonumber \\ \nonumber \\
                               &\leq  \,\, \sigma_{B, max}  (b_1 \epsilon_1 + b_2 \epsilon_2 + b_3 \epsilon_3 + b_4 \epsilon_4 ) 
\label{eq:proof_contractive}
\end{align}
where $\sigma_{B, max}$ stands for the maximum singular value of matrix $\mat{B}$, using the definition of the spectral norm. Note that in inequality (\ref{eq:proof_contractive}), we employed the assumption in (\ref{eq:Lips_deoniser}) that the neural network $f_\theta(\cdot)$ is $\epsilon_1$-Lipschitz continuous, thus the spectral norm of its Jacobian $\partial_xf_\theta(x)$ is bounded by $\epsilon_1$. Additionally, we assume that the convolutional layers $P_1, P_2, P_3$  are $\epsilon_2, \epsilon_3, \epsilon_4$-Lipschitz continuous accordingly.

To further analyze inequality (\ref{eq:proof_contractive}), we need to examine the eigenvalues of the matrix $\mat{B}$. Since $\mat{B} =  (\mathcal{H}^T \mathcal{H} + \alpha \mat{I})^{-1} $, it is easy to prove that the eigenvalues of $\mat{B}$ are related to the eigenvalues of the matrix $\mathcal{H}^T\mathcal{H}$ by the following relation
\begin{equation}
    \lambda_{B,i} = \frac{1}{\alpha + \lambda_{\mathcal{H}^T\mathcal{H}, i} }
    \label{eq:eigen}
\end{equation}
where $\lambda_{\mathcal{H}^T\mathcal{H}, i}$ denotes the i-th eigenvalue of matrix $\mathcal{H}^T \mathcal{H}$. Since the matrix  $\mathcal{H}^T \mathcal{H}$ is symmetric semi-positive definite and the eigenvalues of matrix $\mat{B}$, which are given by equation (\ref{eq:eigen}), are strictly positive,  matrix $\mat{B}$ is positive definite. Thus, based on equation (\ref{eq:eigen}) its singular values  are given by
\begin{equation}
    \sigma_{B,i} = (\lambda_{B,i})^2 =  \frac{1}{(\alpha + \lambda_{\mathcal{H}^T\mathcal{H}, i})^2 }
\end{equation}
In view of this, the maximum singular value of matrix B is 
\begin{equation}
    \sigma_{B,max} = \frac{1}{(\alpha + \lambda_{\mathcal{H}^T\mathcal{H}, min})^2 }
    \label{eq:max_singular}
\end{equation}
where $ \lambda_{\mathcal{H}^T\mathcal{H}, min}$ is the minimum eigenvalue of matrix $\mathcal{H}^T\mathcal{H}$. Based on relation \ref{eq:max_singular}, inequality (\ref{eq:proof_contractive}) can be expressed as
\begin{equation}
  ||\partial_x h_\theta(x; y)|| \leq  \frac{(b_1 \epsilon_1 + b_2 \epsilon_2 + b_3 \epsilon_3 + b_4 \epsilon_4 )}{(\alpha + \lambda_{\mathcal{H}^T\mathcal{H}, min})^2 }
\end{equation}
 This proves that the iteration map $h_\theta(x, y)$ is $\epsilon$-Lipschitz with 
 \begin{equation}
 \epsilon = \frac{(b_1 \epsilon_1 + b_2 \epsilon_2 + b_3 \epsilon_3 + b_4 \epsilon_4 )}{(\alpha + \lambda_{\mathcal{H}^T\mathcal{H}, min})^2 }. 
\end{equation}
Under the above condition, it is ensured that the proposed DeepMix model will converge to a fixed point.


\end{proof}

\section{Experimental part}\label{Results}

In this section, we highlight the effectiveness  of our   method through  experiments that address the hyperspectral deconvolution problem, utilizing both simulated and real-world datasets. 

\subsection{Dataset}
To validate the merits of the proposed DeepMix model, we used three publicly available hyperspectral image datasets, i.e., the CAVE \cite{cave} dataset, the remotely sensed data  Chikusei \cite{yokoya2016airborne} and the real-world dataset from study \cite{sos_pnp}.

\subsubsection{Simulated Experiments}
\begin{itemize}
\item The CAVE dataset consists of 32 hyperspectral images, each with a spatial resolution of 512 by 512 pixels and 31 spectral bands in the range of 400 to 700 nm.
\item The Chikusei dataset is a  hyperspectral image captured by a Visible and Near-Infrared imaging sensor over agricultural and urban regions in Chikusei, Ibaraki, Japan. The dataset consists of $2517 \times 2335$ pixels and covers 128 spectral channels ranging from 363 nm to 1018 nm.
\end{itemize}
All the hyperspectral images were normalized to the range $0-1$. Additionally, following the setup in \cite{wang_main}, we generated degraded images using the following blurring kernels combined with Gaussian noise of standard deviation $\sigma$:
\begin{enumerate}[(a)]
\item $9 \times9$ Gaussian kernel with bandwidth $\sigma_k = 2$, and  $\sigma = 0.01$,
(b) $13 \times 13$ Gaussian kernel with bandwidth $\sigma_k = 3$, and  $\sigma = 0.01$,
(c) $9 \times 9$ Gaussian kernel with bandwidth $\sigma_k = 2$, and $\sigma = 0.03$,
(d) Circle kernel with diameter $7$, and $\sigma = 0.01$, and 
(e) Square kernel with side length of $5$, and $\sigma = 0.01$
\end{enumerate}
In the following experiments, we divided the CAVE dataset into two sets: a training set consisting of the first 20 images, and a test set consisting of the remaining 12 images. For the Chikusei dataset, we extracted a sub-image of size 1024 × 2048 from the top area of the image for training purposes, and then we cropped the remaining part into 32 non-overlapping sub-images of size 256 × 256 × 128 for testing.

\subsubsection{Real-world experiments}
Finally, in Section \ref{real-world}, \textit{we  showcase the effectiveness of our  method in real-world scenarios. These scenarios are characterized by more intricate blurring effects, with blur kernels that vary across spectral bands and are unknown, thus requiring estimation. For this purpose, we have chosen to utilize a real dataset from  \cite{sos_pnp}, which consists of unfocused hyperspectral images.}

\subsection{Implementation Details}

\subsubsection{Hyperparameters setting}
Regarding the DeepMix model, the Anderson acceleration technique \cite{aderson} was employed to speed up the fixed-point computations during the forward and backward passes in both training and testing phases, with the number of iterations set to $15$. In the end-to-end training process, the Adam optimizer was used with a learning rate of $1e-04$ and a batch size of $6$ for $15$ epochs. 

\begin{figure*}
\centering
\setlength{\tabcolsep}{0.01pt} 
\begin{tabular}{ccccc}
\includegraphics[width=0.15\linewidth]{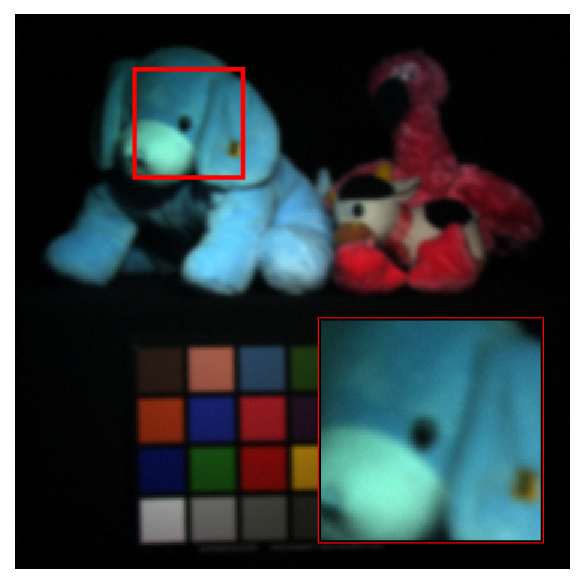} &
\includegraphics[width=0.15\linewidth]{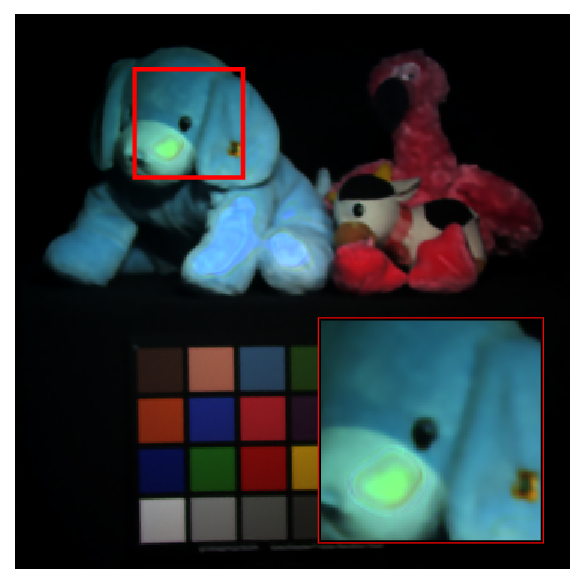} &
\includegraphics[width=0.15\linewidth]{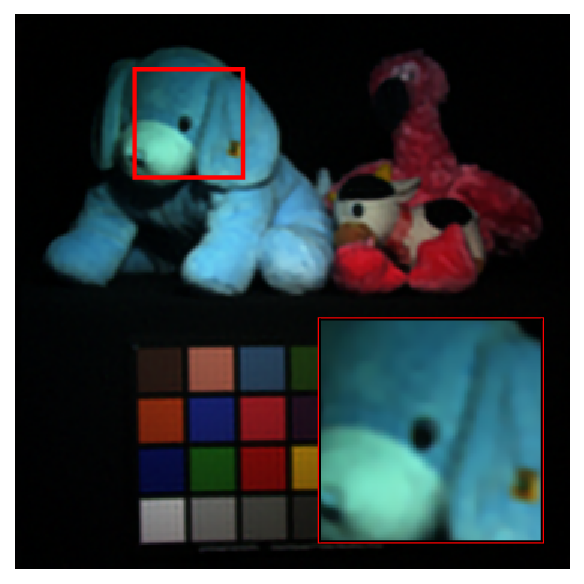} &
\includegraphics[width=0.15\linewidth]{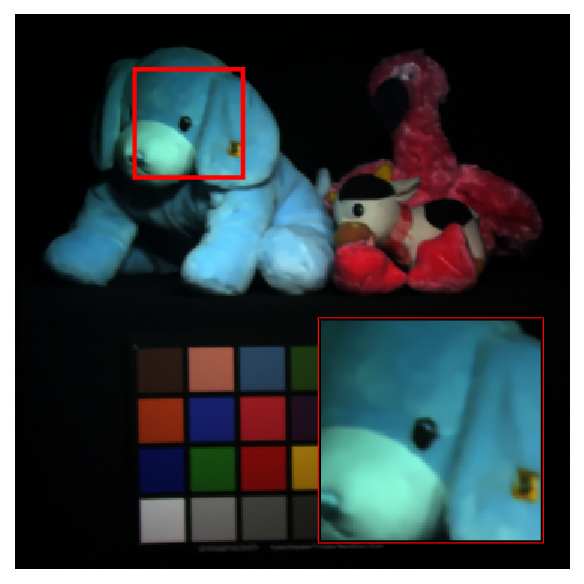} &
\includegraphics[width=0.15\linewidth]{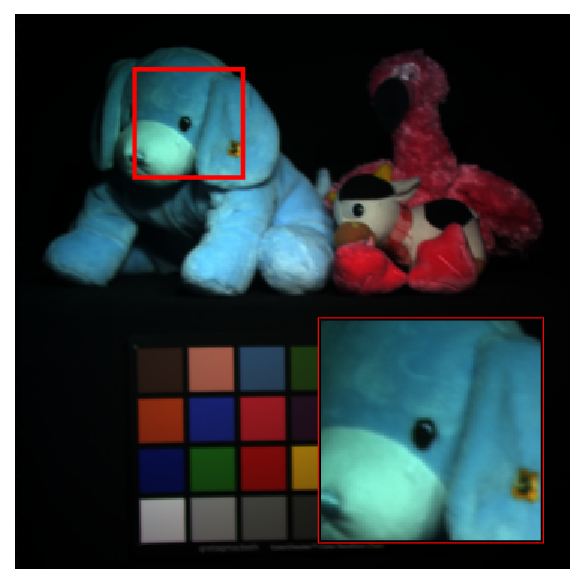} \\
\includegraphics[width=0.15\linewidth]{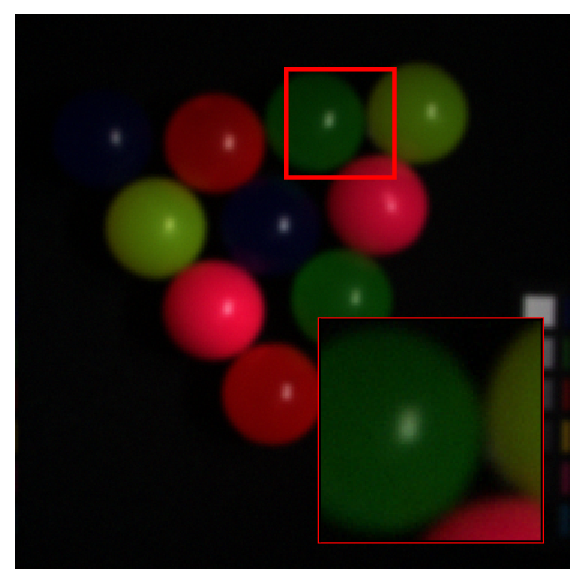} &
\includegraphics[width=0.15\linewidth]{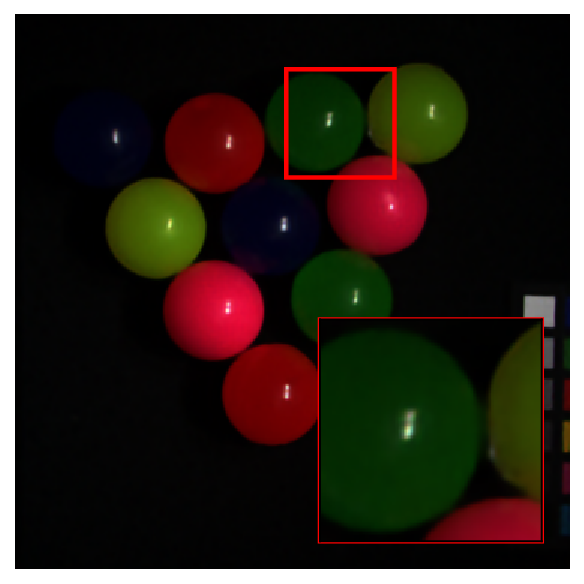} &
\includegraphics[width=0.15\linewidth]{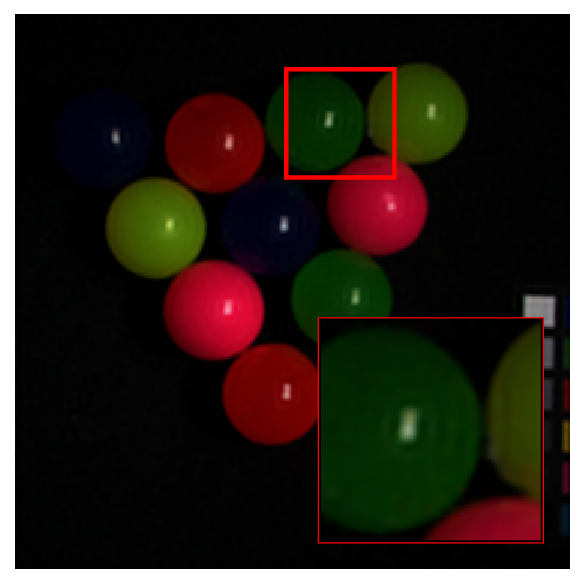} &
\includegraphics[width=0.15\linewidth]{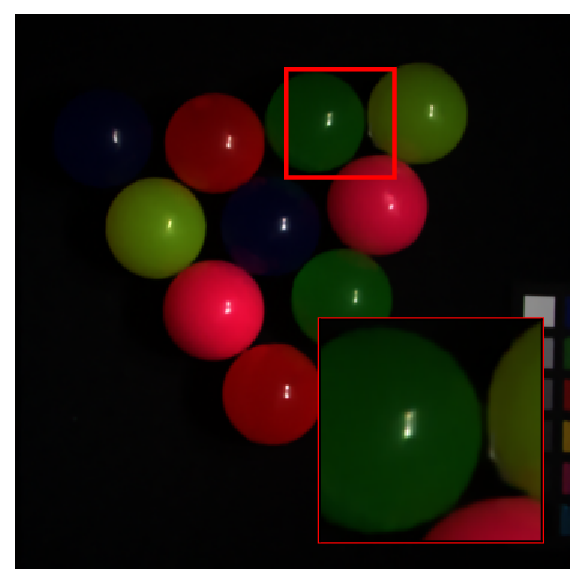} &
\includegraphics[width=0.15\linewidth]{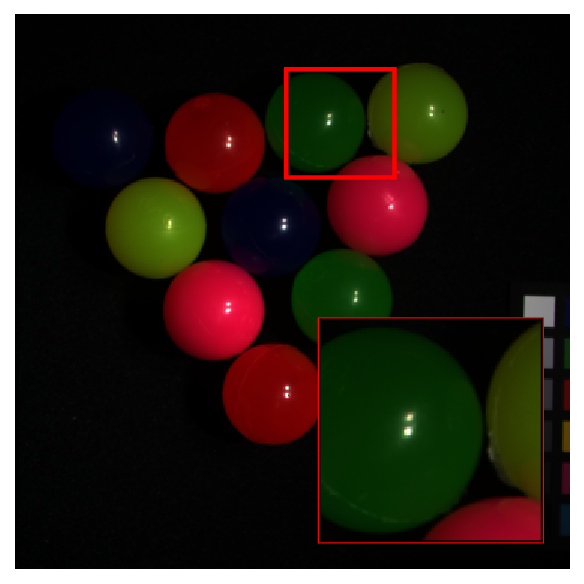} \\
\multicolumn{1}{c}{(a) Blurred Image} & 
\multicolumn{1}{c}{(b) Hsi-denet \cite{8435923}} & 
\multicolumn{1}{c}{(c) PnP \cite{sos_pnp}} & 
\multicolumn{1}{c}{(d) DeepMix} & 
\multicolumn{1}{c}{(e) Ground Truth}
\end{tabular}
\caption{Visual results for the best performing methods in the blurring scenario (a) on the CAVE dataset. The ﬁrst and second rows present the results for two different blurred
images. The false color images were generated for clear visualization with the 22th, 14th and 7th channels used for red, green and blue.}
\label{fig:cave_fig}
\end{figure*}

\subsubsection{Quantitative metrics} For the evaluation, we used the following metrics:
(i) Peak Signal to Noise Ratio (PSNR)
(ii) Structural Similarity Index (SSIM)\cite{sim}
(iii) Root Mean Square Error (RMSE) and (iv)
Erreur Relative Globale Adimensionnelle de Synthese (ERGAS) \cite{ergas}

\subsubsection{Compared Methods}

a) Optimization-based methods:
We considered several optimization-based methods that employ either handcrafted or learnable regularizers. Notable examples of these methodologies include Hyper-Laplacian Priors (HLP)  \cite{method_2}, Spatial and Spectral Priors (SSP) \cite{method_3}, Weighted Low-Rank Tensor Recovery (WLRTR) \cite{wltr}, 3D Fractional Total Variation (3DFTV) \cite{tv}. Additionally, utilizing the open-source HSI toolbox Hyde \cite{Hyde}, we  considered other three  optimization based restoration approaches, i.e., the method L1HyMixDe \cite{l1hyde}, the approach Hyres \cite{hyres} and the method FORDN \cite{hyres}. b) Deep learning methods:   the SOTA  method \cite{sos_pnp}, and the  Hsi-denet \cite{8435923} approach.

\subsection{Performance evaluation on simulated data}
\begin{figure*}
\centering
\setlength{\tabcolsep}{0.005pt}
\begin{tabular}{ccccc}
\includegraphics[width=0.15\linewidth]{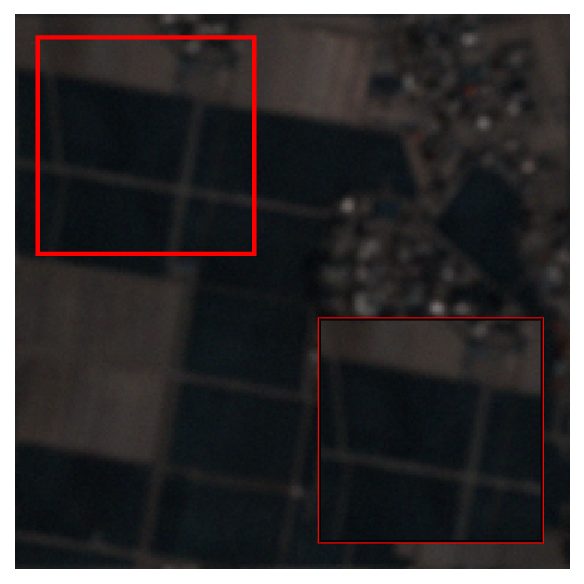} &
\includegraphics[width=0.15\linewidth]{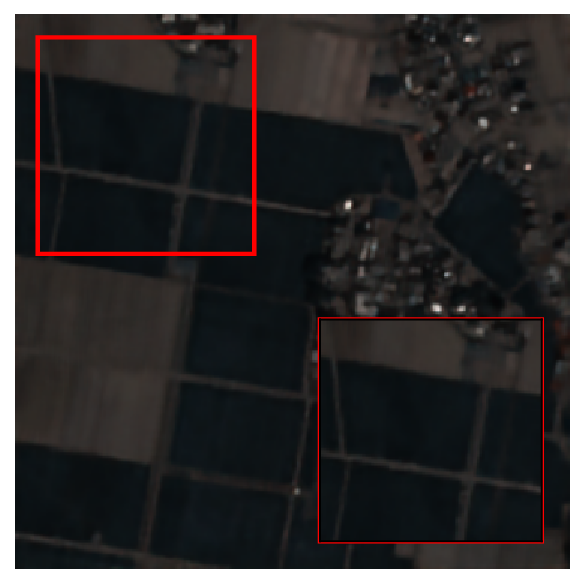} &
\includegraphics[width=0.15\linewidth]{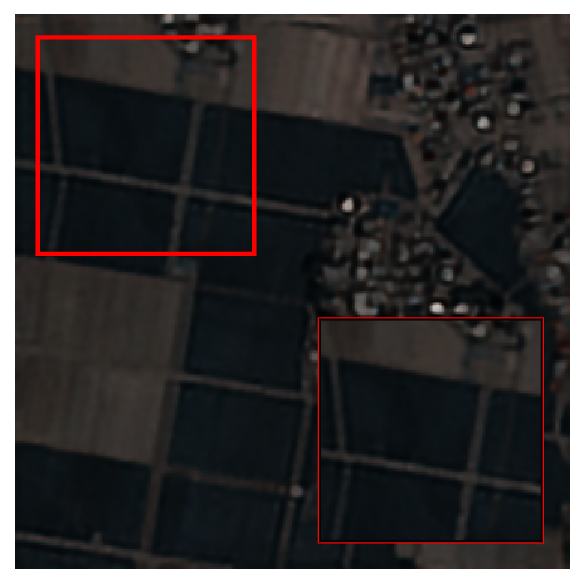} &
\includegraphics[width=0.15\linewidth]{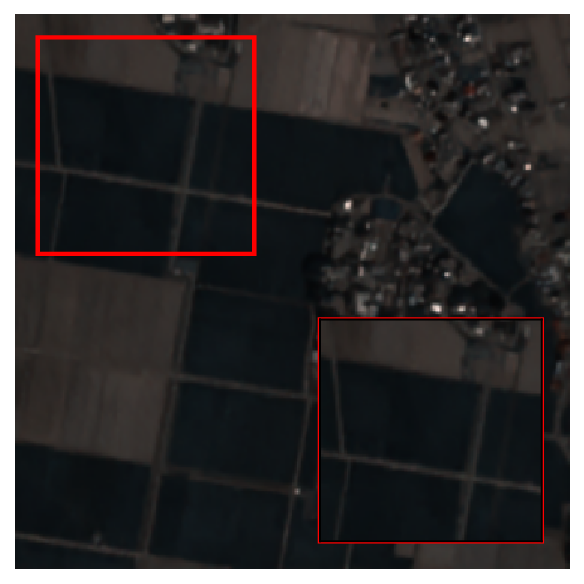} &
\includegraphics[width=0.15\linewidth]{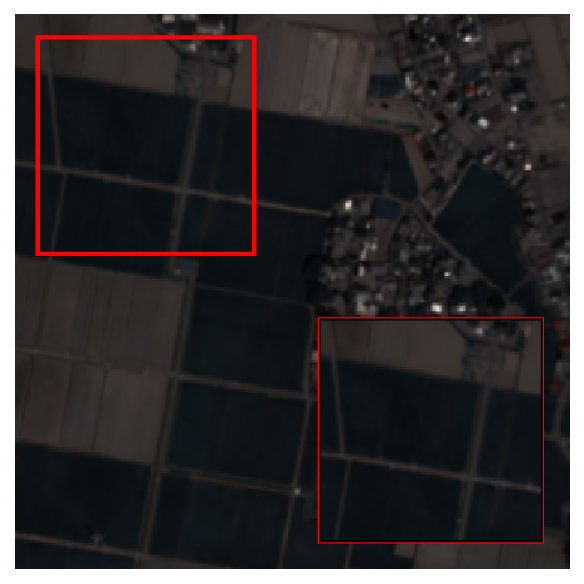} \\
(a) Blurred Image & (b) Hsi-denet \cite{8435923} & (c) PnP \cite{sos_pnp} & (d) DeepMix& (e) Ground Truth  
\end{tabular}
\caption{Visual results for the best performing methods in the blurring scenario (b) on the Chikushei dataset. The false color images were generated for clear visualization with the 122th, 84th and 57th used for red, green and blue, respectively.}
\label{fig:chikushei-fig}
\end{figure*}

In this section, we validated the HSI deconvolution performance of the proposed DeepMix model against several  other approaches using the CAVE and remotely sensed Chikusei datasets. 
Tables \ref{tab:cave} and \ref{tab:chikusei} summarize the average results in terms of RMSE, PSNR, SSIM and ERGAS under various blurring scenarios. It is evident that the proposed DeepMix model consistently outperforms the other methods in all scenarios for both datasets.  Furthermore, the proposed DeepMix model demonstrates superior restoration performance, particularly when dealing with high levels of noise as   compared to
the other methods, whose performance degrades significantly for high levels of noise. 
The above superior performance of the proposed model is further validated from Figures \ref{fig:cave_fig} and \ref{fig:chikushei-fig}, where the DeepMix model is shown to be able to provide  clearer and sharper images as compared to the two best performing methods (study \cite{sos_pnp} and study \cite{8435923}.)


\textbf{Comparison with the deep learning Plug-and-play (PnP) method}: Compared to the  PnP approach in \cite{sos_pnp}, the proposed methodology provides notably better reconstructions results. The superior performance of the proposed DeepMix model can be attributed to the end-to-end optimization of its learnable parameters. These components are optimized considering the quality of the reconstructed images and are specifically adapted to the examined HSI deconvolution problem. In contrast, the PnP method does not utilize an end-to-end training. The denoiser prior is used by the PnP method separately from both the HSI degradation problem and the blurring operator.
\textbf{Comparison with the deep learning method \cite{8435923}}: Compared to the conventional deep learning method outlined in \cite{8435923}, our model exhibits  significant benefits. The traditional deep learning framework involves over $1$ million learnable parameters, demanding extensive data for effective training. In scenarios where the data are limited as those in our case, our model is designed to be significantly more efficient, containing $98\%$ fewer parameters than the deep learning method a property derived from its optimization-based nature.

\textbf{Runtime - real-time application:}
In terms of runtime,  Table \ref{tab:runtime} demonstrates the average execution times of the evaluated methods during the testing phase. Remarkably, the proposed DeepMix method not only achieves improved results but also exhibits considerably faster hyperspectral image reconstruction times as compared to the optimization-based and the plug-and-play methods. 

\begin{table*}
\centering
  \caption{Proposed method versus several approaches: Results under several blur scenarios on CAVE dataset.}
  {
  \resizebox{12cm}{!}{
  \begin{tabular}{cc|ccccccccccl}
    \toprule
    Blurring& Metrics&L1HyMixDe& Hyres& FORDN & HLP  &SSP  & WLRTR  & 3DFTV         & PnP  &Hsi-denet  & proposed  \\
    Scenarios&          & \cite{l1hyde} &\cite{fordn} &\cite{fordn}  & \cite{method_2}   &\cite{method_3} & \cite{wltr} & \cite{tv}  & \cite{sos_pnp} & \cite{8435923}  &  DeepMix \\ 
    
    \midrule
         &RMSE   &4.323 &5.94& 5.61            & 4.420  & 4.848  &4.735 &4.332    & 3.132                    & 4.61                          &\textbf{2.5367}    \\
      (a) &PSNR  & 35.94& 35.02& 34.98           & 36.166  & 35.373 &35.872  &36.450 &39.252                 & 37.89                         &  \textbf{41.51}    \\
         &SSIM   &0.9569 &0.9199 &0.9282            &  0.9167 & 0.9305 & 0.9380  &0.9401 & 0.9493            &0.9480                         & \textbf{0.9772}    \\
         &ERGAS  & 18.986 &20.89 &20.71           &  18.15&19.51  &18.96 &17.34  &13.01                      & 13.86                          & \textbf{9.378}   \\
    \midrule
         &RMSE   &5.479 &6.334 &  5.379          &  5.707& 5.955 & 6.439 &5.667& 4.581                       & 4.134                         & \textbf{3.395}   \\
      (b) &PSNR  & 33.14&32.57 &   33.21         & 34.034 & 33.541  & 33.084 & 34.116& 36.305                & 36.78                          &  \textbf{38.68}    \\
         &SSIM   & 0.8598& 0.8223&  0.8389          &  0.8911  &0.9031 & 0.9025& 0.9136&0.9234               & 0.9478                          & \textbf{0.9576}    \\
         &ERGAS  & 22.74&24.43 &    22.40        & 22.92  &23.71 &25.46 &22.40 &18.54                        & 14.73                          & \textbf{12.33}    \\
    \midrule
         &RMSE   & 5.242 &6.48 &5.97           & 7.669  &5.270 & 5.099 & 5.016 & 4.225                       & 4.275                    & \textbf{3.338}     \\
      (c) &PSNR  & 34.50 &32.05 &33.34           &30.599 & 34.309  &34.827 &34.741&36.211                    & 36.15                    &   \textbf{38.51}    \\
         &SSIM   & 0.8752 &0.7964 &0.8916            &  0.6406  & 0.8565 &0.8956  &0.8851  &0.8708           & 0.8698                     & \textbf{0.9486}     \\
         &ERGAS  & 21.41 &26.17 &22.3           &  33.49& 22.28 & 20.80 &20.47& 18.64                        & 16.78                      & \textbf{12.23}    \\
    \midrule
         &RMSE   &5.054  &5.43 &4.968             & 4.189 & 4.584 & 4.328       &4.167     &2.305            & 2.8105                  & \textbf{1.705}  \\
      (d) &PSNR  & 35.12 &34.22 &35.171            & 36.549 & 35.862   &36.68   &36.80     &41.65            & 40.56                    & \textbf{44.50}      \\
         &SSIM   & 0.9468 &0.9214 & 0.9390            &  0.9165&0.9354 & 0.9450  &0.9403       &0.9542       & 0.9666                     & \textbf{0.9754}    \\
         &ERGAS  & 18.61 &20.17 & 18.50            & 17.36 & 18.49&17.45       &16.69        &9.86           &10.14                         &  \textbf{6.430} \\
         
    \midrule
         &RMSE  & 4.890 &5.285  &5.133              & 3.971 & 4.356 & 4.109  &3.957  &2.280                  & 2.298                        & \textbf{1.623}     \\
      (e) &PSNR & 35.41 &34.45  &34.89              & 36.910 & 36.322&37.130 &37.225  &41.932                & 41.89                     &  \textbf{44.87}    \\
         &SSIM             & 0.9501 & 0.9240 & 0.9358            &  0.9195 & 0.9397  &0.9480& 0.9468 &0.9475           & 0.9722                      &\textbf{0.97821}   \\
         &ERGAS            & 18.09 & 19.71 &19.03            &  16.58 & 17.60 &16.64 &15.89 &9.79                      & 8.25                           & \textbf{6.11}   \\
    \bottomrule
  \end{tabular}}}
    \label{tab:cave}
\end{table*}

\subsection{Ablation Analysis}
\subsubsection{Interplay between Handcrafted and Learnable Regularizers: Effects on Denoiser Complexity} \label{handcrafted_and_learnable}

\begin{table*}
\centering
  \caption{Proposed method versus several approaches: Results under several blur scenarios on Chikusei dataset.}
  {
  \resizebox{11cm}{!}{
  \begin{tabular}{cc|ccccccccccl}
    \toprule
    Blurring& Metrics&L1HyMixDe& Hyres& FORDN & HLP  &SSP  & WLRTR  & 3DFTV         & PnP  &Hsi-denet  & proposed  \\
    scenarios&          & \cite{l1hyde} &\cite{fordn} &\cite{fordn}  & \cite{method_2}   &\cite{method_3} & \cite{wltr} & \cite{tv}  & \cite{sos_pnp} & \cite{8435923} &DeepMix \\ 
    
    \midrule

        &RMSE    & 4.27  &4.41   &  4.321          & 3.233 & 3.050& 3.138 & 3.207 & 2.560 &          2.898         & \textbf{1.328}\\
   (a)     &PSNR & 36.99  &36.21   &  36.32           &38.979 & 40.182 &40.051 & 39.546  &41.032 &     39.45         & \textbf{41.67}\\
        &SSIM    & 0.9116  &0.8943   & 0.9014             & 0.9124 & 0.9334 &0.9267 & 0.9171 & 0.9420  &   0.9587        & \textbf{0.9798}\\
        &ERGAS   & 31.72  & 37.89  &  36.87           &32.25 & 28.13 & 25.29 &35.37  &27.87&              20.83      &\textbf{15.33} \\ 
      \midrule
        &RMSE    & 5.378  & 5.50  &   5.412          & 3.945 & 3.819 & 4.091  &4.037& 3.428 &        4.112           & \textbf{2.301}\\
   (b)     &PSNR &  34.98 &34.86   &    34.02          & 37.604& 38.392 & 37.872 & 37.708 & 38.989 &   36.95         &\textbf{39.94} \\
        &SSIM    & 0.8621  &0.8564   &   0.8612          &0.8822 & 0.9016 &0.8871 &0.8819 & 0.9091  &  0.9101        &\textbf{0.9648} \\
        &ERGAS    & 39.11 &40.35   &   39.50         &35.30  &32.40 &31.45 &39.85 &30.92 &             27.14         &\textbf{17.85} \\
      \midrule   
        &RMSE     & 4.444 & 5.30  &  4.69          & 7.094& 3.506 & 3.777& 3.662  &3.413  &       3.541              & \textbf{2.834}\\
   (c)     &PSNR  & 35.89 & 33.95  &  35.13          & 31.391& 37.942  &37.447& 37.756 &37.934&   37.87              &\textbf{39.60}\\
        &SSIM     & 0.8799 & 0.8126  &  0.8792          & 0.6268 & 0.8839 & 0.8816 & 0.8841 &0.8783   &0.9098        &\textbf{0.9499}\\
        &ERGAS    &52.24  &59.33   &   55.52         &90.14 & 50.26& 39.95 &48.15 & 51.38 &           36.46          &\textbf{23.23}  \\
        \midrule
        &RMSE     & 4.44 & 4.21  &   4.102         &3.361 & 2.879 & 2.890 & 3.076  &2.335              &2.324         &\textbf{1.759} \\
    (d)    &PSNR  &36.75  & 36.89  &   36.74          & 39.122& 40.625 &40.724 & 39.900 &41.290        &41.35         & \textbf{43.74}\\
        &SSIM     & 0.8975 & 0.9005  & 0.9012            &0.9148 & 0.9399 & 0.9364 & 0.9228 & 0.9430   &0.9678        &\textbf{0.9771} \\
        &ERGAS    & 36.81 & 34.78  &   34.87         &32.76 & 27.22  &23.73& 34.59 & 32.56             &27.58         & \textbf{15.51}\\
     \midrule
        &RMSE      & 4.025 &4.014  &  3.901        &2.990& 2.688 & 2.691  &2.913 & 2.148               &2.105         & \textbf{1.548}\\
    (e) &PSNR     & 36.99  & 37.01 & 37.10            & 39.352 & 41.174 & 41.313 & 40.334 & 41.971     &42.01         & \textbf{44.13}\\
        &SSIM      & 0.9193 & 0.9090 & 0.9192         &0.9188& 0.9456  & 0.9438  & 0.9295 &  0.9506    &0.9665        & \textbf{0.9812}\\
        &ERGAS     & 32.00 &31.02  &  32.15        & 32.68 &26.19 & 22.46 & 33.74  &30.62              &17.84          & \textbf{14.28}\\ 
    \bottomrule
  \end{tabular}}}
    \label{tab:chikusei}
\end{table*}

Recall that our  DeepMix model comprises three interpretable components: the data consistency module (Eq. \ref{eq:x}), the context aware denoising module (Eq. \ref{eq:denoiser}), and the smooth gradient priors module, outlined by equations \ref{eq:gradient_1}, \ref{eq:gradient_2}, and \ref{eq:gradient_3}. In this experiment, using the CAVE dataset, our aim is to examine the effect of the incorporation  of handcrafted regularizers on the complexity of the context aware denoising module.
 Specifically, we compare our model with a "simpler" variant that relies solely on a learnable regularizer, thus only including the data consistency and context aware denoising modules. To assess the models' complexity, both configurations of the denoiser were tested with varying numbers of 3D convolutional blocks, namely $N=2, 3$, and $4$ (refer to Figure \ref{fig:3d-cnn} for more  details). The findings are summarized in Table \ref{tab:psnr_handcrafted}.

\textbf{Benefits on restoration performance:} The results illustrate that the integration of handcrafted and learnable regularizers significantly enhances the performance of our  model in all blurring scenarios, compared to the approach that solely utilizes the learnable regularizer.
\textbf{Benefits on complexity:} In addition to the substantial improvement in performance, this integration allows the denoiser to be more compact and computationally efficient. This is particularly evident as our DeepMix model, requires a denoiser with only $N=2$ convolutional blocks to surpass the performance of the other considered model  that requires a more complex denoiser with $N=4$ convolutional blocks,  resulting in a model with $\bold{50}\%$ fewer learnable parameters.
This notable improvement can be attributed to the handcrafted regularizers, which capture a significant portion of the hyperspectral images' structure, thereby minimizing the "residual structure" that the denoiser needs to address, thus requiring a simpler neural network to capture it. Conversely, when a model relies solely on the denoiser for capturing the structure of the data, it necessitates a higher complexity.  These findings highlight the value of combining handcrafted priors with deep learning priors, leading to both higher accuracy and greater computational efficiency.
\begin{table}
  \caption{Average runtime (in seconds) of the considered methods .}
    \begin{center}
    \resizebox{\linewidth}{!}{
    \begin{tabular}{cccccccc}
    \toprule
         Method & HLP  & SSP& WLRTR  & 3DFTV & PnP & Hsi-denet & proposed DeepMix\\

        \midrule
        time[sec] & 9.7 & 622.5 & 10501.2 & 6044.4 & 4280.6 & 3.7& 4.1  \\
        \bottomrule
    \end{tabular} }
    \end{center}
    \label{tab:runtime}
    
\end{table}

\begin{figure*} 
\centering
\setlength{\tabcolsep}{0.01pt} 
\begin{tabular}{ccccc}

\includegraphics[width=0.15\linewidth]{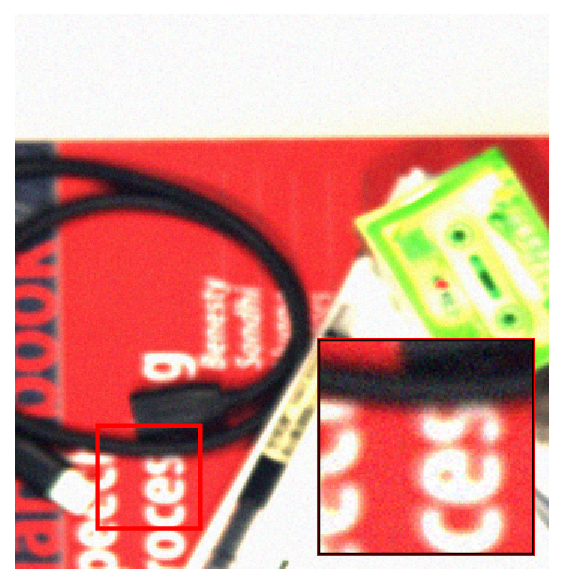} &
\includegraphics[width=0.15\linewidth]{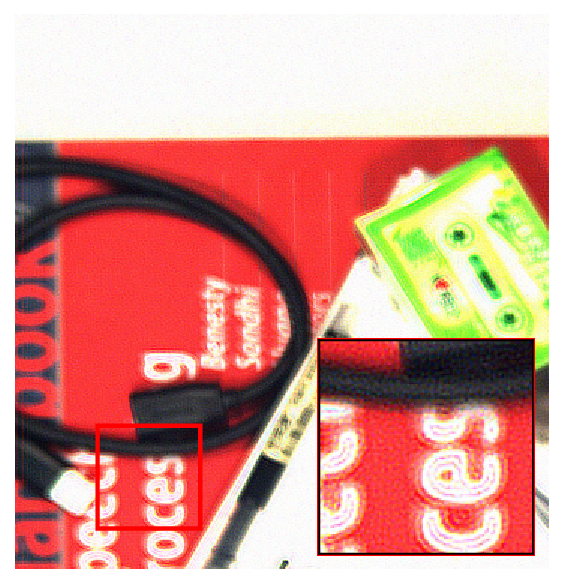} &
\includegraphics[width=0.15\linewidth]{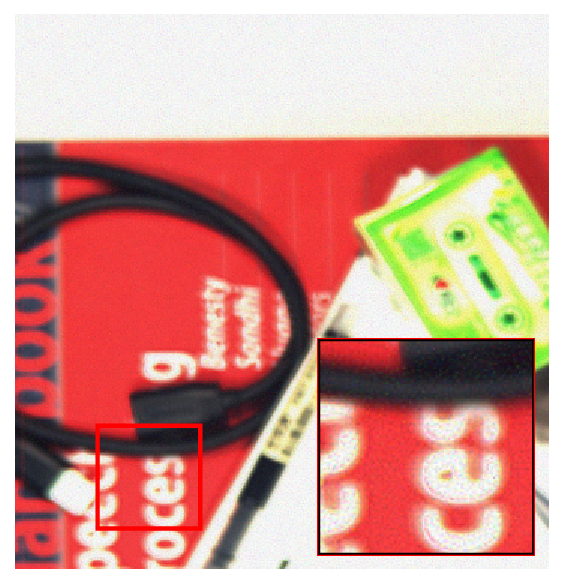} &
\includegraphics[width=0.15\linewidth]{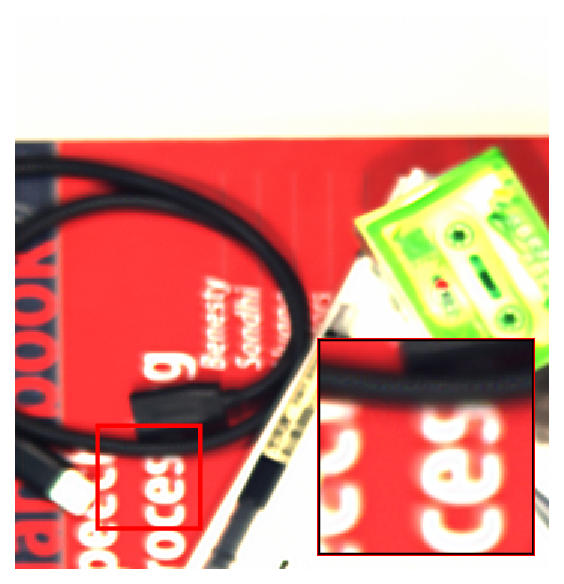} &
\includegraphics[width=0.15\linewidth]{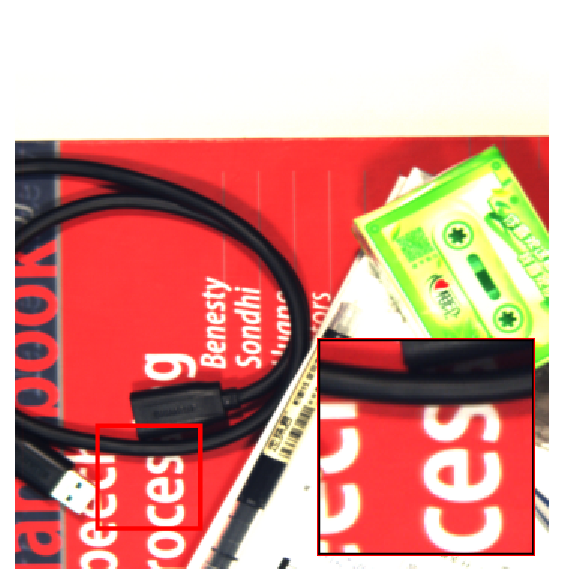} \\
\includegraphics[width=0.15\linewidth]{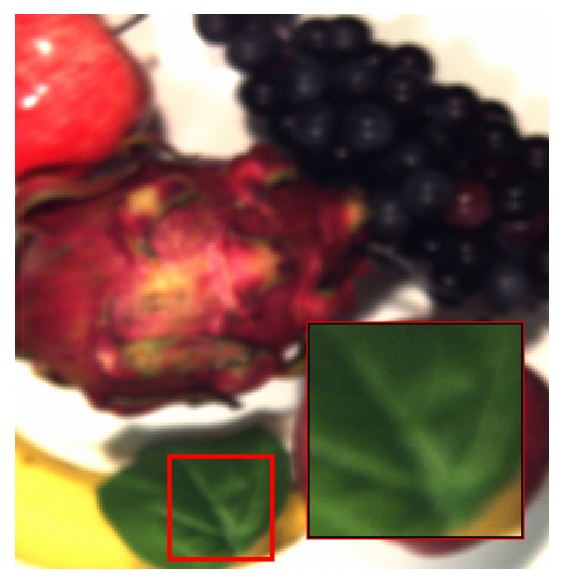} &
\includegraphics[width=0.15\linewidth]{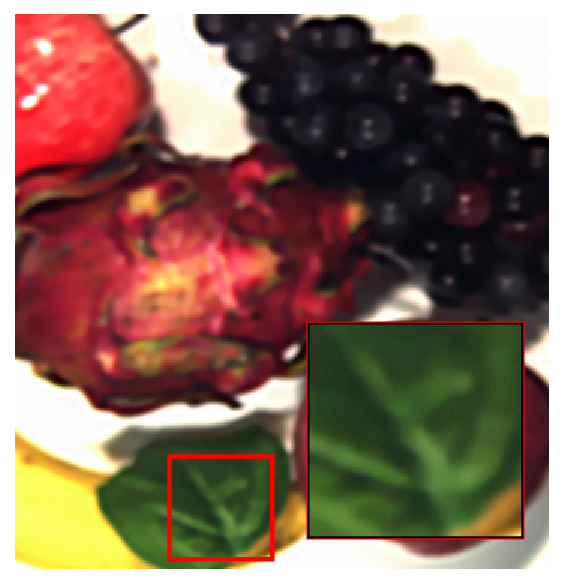} &
\includegraphics[width=0.15\linewidth]{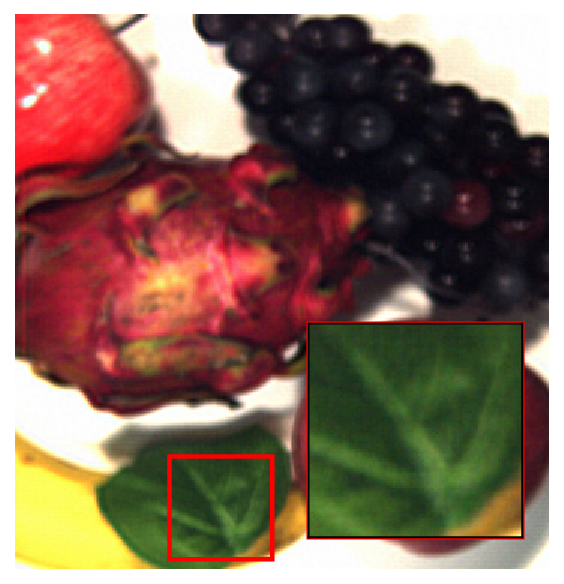 } &
\includegraphics[width=0.15\linewidth]{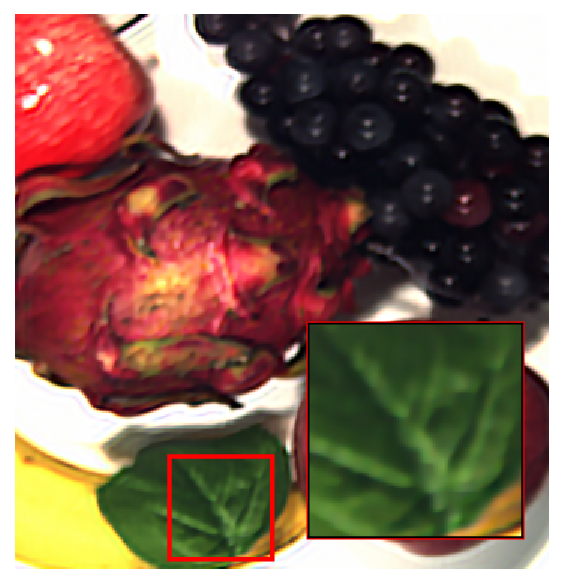} &
\includegraphics[width=0.15\linewidth]{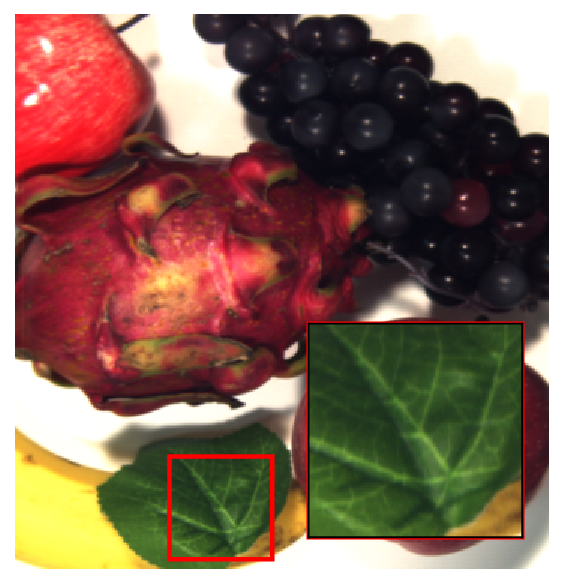} \\
\multicolumn{1}{c}{(a) Blurred Image} & 
\multicolumn{1}{c}{(b) Hsi-denet \cite{8435923}} & 
\multicolumn{1}{c}{(c) PnP \cite{sos_pnp}} & 
\multicolumn{1}{c}{(d) DeepMix} & 
\multicolumn{1}{c}{(e) Ground Truth}

\end{tabular}
\caption{Blurred images, reference images and visual results for the best performing methods on the real-world dataset. The false color images were generated for clear
visualization with the 38th, 24th and 10th channels used for red, green and blue, respectively.}
\label{fig:real-data}
\end{figure*}

\begin{table}
\centering
\caption{PSNR comparison with and without handcrafted regularizers}
\label{tab:psnr_handcrafted}
\resizebox{7cm}{!}{
\begin{tabular}{lccc}
\toprule
Blurring & Complexity of Denoiser& No Handcrafted & proposed \\
Scenarios  &    No. of Blocks & only denoiser &                        \\
\midrule
                                                 & 2 & 40.74 & 41.13 \\
(a)& 3 & 40.93 &  41.29\\
                                                 & 4 &  41.12& 41.51 \\
\midrule
 & 2 & 38.28 & 38.70\\
(b)  & 3 & 38.51 & 38.88 \\
 & 4 & 38.67 & 38.99 \\
\midrule
 & 2 & 37.81 & 38.23 \\
(c)  & 3 & 38.10 & 38.43 \\
 & 4 & 38.19 & 38.50 \\
\midrule
 & 2 & 42.34 & 43.90 \\
(d)  & 3 & 43.74 & 44.17 \\
 & 4 & 43.85 & 44.50 \\
\midrule
& 2 & 43.01 & 44.32 \\
(e)  & 3 & 43.49 & 44.55 \\
& 4 & 43.36 & 44.87 \\
\bottomrule
\end{tabular}}
\end{table}

\begin{table*}
\centering
\caption{The impact of skip connection in the performance of the proposed DeepMix: PSNR comparison  with and without skip connections in the CAVE and Chikushei datasets}
\label{tab:psnr_skip}
\resizebox{11cm}{!}{
\begin{tabular}{lccc||cc}
\toprule
Type of Kernel & \multicolumn{1}{c}{No. of Blocks} & \multicolumn{2}{c||}{CAVE Dataset} & \multicolumn{2}{c}{Chikusei} \\
\cline{3-6} 
               &  & \multicolumn{2}{c||}{Denoiser} & \multicolumn{2}{c}{Denoiser} \\
               &  & No skip-connection & With skip-connection & No skip-connection & With skip-connection \\
\midrule
(a) & 2 & 38.77 & 41.13 & 36.95 & 41.12 \\
    & 3 & 40.37 & 41.29 & 37.55 & 41.36 \\
    & 4 & 40.58 & 41.51 & 38.45 & 41.67 \\
\midrule    
(b) & 2 & 36.98 & 38.70 & 36.01 & 39.21 \\
    & 3 & 37.77 & 38.88 & 36.87 & 39.85 \\
    & 4 & 38.05 & 38.99 & 37.06 & 40.32 \\
\midrule
(c) & 2 & 36.61 & 38.23 & 35.90 & 38.85 \\
    & 3 & 37.51 & 38.43 & 36.32 & 39.15 \\
    & 4 & 37.84 & 38.50 & 37.08 & 39.60 \\
\midrule
(d) & 2 & 42.01 & 43.90 & 39.99 & 42.95 \\
    & 3 & 42.59 & 44.17 & 40.70 & 43.37\\
    & 4 & 43.30 & 44.50 & 41.01 & 43.74 \\
\midrule
(e) & 2 & 43.01 & 44.32 & 40.02 & 43.25 \\
    & 3 & 43.49 & 44.55 & 40.87 & 43.87 \\
    & 4 & 43.70 & 44.87 & 41.12 & 44.13 \\
\bottomrule
\end{tabular}}
\end{table*}

\subsubsection{The impact of skip connection in the performance of the proposed DeepMix} \label{skip_connection}

In this experiment, the impact of the skip-connection on the architecture of the context aware denoiser  is evaluated using the CAVE and Chikushei datasets utilizing various blurring scenarios. 
In particular, we considered two configurations of our model: one incorporating the skip connection  and a second model that employs a denoiser without the skip connection. The significant benefits of the skip-connection are summarized in Table \ref{tab:psnr_skip}. From this table, it can be seen that the proposed scheme notably outperforms the model which does not consider the skip-connection in every blurring case and across both datasets.  Notably, the skip connection not only enhances restoration quality but also contributes to computational efficiency. For instance, our DeepMix model with the skip connection achieves superior performance containing a denoiser with just 2 convolutional blocks, in contrast to the model without the skip connection, which requires a more complex denoiser with 4 convolutional blocks namely, resulting in a model with $\bold{50}\%$ fewer learnable parameters. 
This finding strongly supports the critical role of a properly integrated skip connection within our context aware denoising module. By facilitating more effective information flow among the collaborating components of our DeepMix model, it ensures that essential details and structures, initially identified by the data-consistency and smooth prior modules, are preserved and not lost during the denoising process.

\subsubsection{Uncertainty to blurring kernel - Generalizability} \label{gener}

\begin{table}
\centering
\caption{PSNR results for the CAVE dataset using the  DeepMix model trained with blurring kernel (a) and tested with various blurring kernels, compared to the  PnP method}
\label{tab:psnr_cave}
\resizebox{8cm}{!}{
\begin{tabular}{l|c|c}
\hline
\textbf{Test Blurring Kernel} & \textbf{ DeepMix(dB)} & \textbf{PnP \cite{sos_pnp} (dB)} \\ \hline
(a) 9x9 Gaussian,    $\sigma_k$=2, $\sigma$=0.01 & 41.51 & 39.252 \\
(b) 13x13 Gaussian, $\sigma_k$=3, $\sigma$=0.01 & 38.11 & 36.305 \\
(c) 9x9 Gaussian, $\sigma_k$=2, $\sigma$=0.03 & 38.92 & 36.211 \\
(d) Circle, diameter 7, $\sigma$=0.01 & 43.34 & 41.65 \\
(e) Square, side length 5, $\sigma$=0.01 & 43.99 & 41.93 \\ \hline
\end{tabular}}
\end{table}

In this experiment, our objective is to evaluate the generalizability of the proposed DeepMix. This is an important aspect since real-world applications often involve unknown blurring kernels, which can only be estimated, thus resulting in using noisy approximations to the actual kernel. 
Consequently, in many practical cases, the model may be trained with a specific blurring kernel, while during the inference phase, the actual kernel may be different.
Taking this into account, we trained the DeepMix model using the CAVE dataset with the blurring kernel in scenario (a), and during the testing phase, we employed several different blurring kernels, which are listed in Table \ref{tab:psnr_cave}. Additionally, we compared the proposed DeepMix approach against the state-of-the-art Plug-and-Play (PnP) method. 
As observed in Table \ref{tab:psnr_cave}, the DeepMix model demonstrates strong generalization properties, as also in this case, it outperforms the PnP method across various blurring kernels of different sizes and types, which are distinct from the one used during training.
In conclusion, the proposed  architecture exhibits strong generalization properties, making it a promising learning-based approach for real-world applications. This important finding is further validated in Section \ref{real-world} in a real-world scenario, where the blurring kernel is unknown and the blurring effects are more complex.

\subsection{Performance evaluation on real-world data} \label{real-world}
To assess the real-world applicability of our method, we utilized a dataset from the study referenced as \cite{sos_pnp}. This dataset includes both unfocused hyperspectral images and their focused counterparts, curated to specifically address the challenges of hyperspectral deconvolution due to defocusing effects. The unfocused images were intentionally created by adjusting the camera's focus, while the focused images provide a clear reference for comparison. The dataset features images with a spatial resolution of 780x696 pixels and with spectral range of 370 to 1000 nm. A notable point is the potential misalignment between image pairs, caused by variations in camera positioning. Therefore, the focused images were utilized solely for qualitative visual comparisons.
Furthermore, the dataset encompasses a blurring kernel unique to each spectral band, estimated using the method cited in \cite{5995521}. Illustrations of the blurred images and their corresponding ground truths are presented in the first and fifth columns of Figure \ref{fig:real-data}, respectively.

A significant challenge in our experiment was the unknown blurring kernel and the absence of a ground-truth image for model training. Consequently, we applied a model previously trained on the CAVE dataset, which features a distinctly different kernel and spectral bands. The visual comparisons in Figure \ref{fig:real-data} clearly demonstrate that our  DeepMix model achieves superior clarity over the current state-of-the-art methods cited in \cite{sos_pnp} and \cite{8435923}. This observation underscores the robustness and generalization capability of our method in practical applications, further supported by its reported low computational complexity in Table \ref{tab:runtime}.

\section{Conclusions}\label{conclusions}
Inspired by the deep unrolling modeling, we have designed an end-to-end trainable deep network that combines the flexibility of optimization-based methods, such as generalizability and convergence properties, with the benefits of learning-based approaches, e.g. superior restoration performance and computation time efficiency.  Specifically, the proposed model comprises three interpretable modules: the data consistency step, the prior denoiser module and the smooth prior module, ensuring that the network can effectively combine the structure of the degradation model  and a-priori knowledge about hyperspectral images. The integration and interaction of these modules are notably enhanced through the use of strategically placed skip connections, facilitating improved preservation of critical image features and reducing redundant modeling.
Extensive experimental results demonstrate the flexibility, and generalizability of our method not only in various simulated blurring settings but can also in real-world scenarios.  In the future, we aim to tackle the blind deconvolution problem, utilizing the deep unrolling framework and incorporating the learning of the kernel into the optimization problem.

\bibliographystyle{IEEEtran}
\bibliography{mybibfile}

\end{document}